\documentclass[twoside]{article}

\usepackage[preprint]{aistats2027}
\usepackage[round]{natbib}

\usepackage[utf8]{inputenc} % allow utf-8 input
\usepackage[T1]{fontenc}    % use 8-bit T1 fonts
\usepackage{url}            % simple URL typesetting
\usepackage{booktabs}       % professional-quality tables
\usepackage{amsfonts}       % blackboard math symbols
\usepackage{nicefrac}       % compact symbols for 1/2, etc.
\usepackage{microtype}      % microtypography
\usepackage{xcolor}         % colors
\usepackage{graphicx}
\usepackage{subcaption}
\usepackage{booktabs} % for professional tables
\usepackage{tikz}
\usepackage{circuitikz}
\usepackage{algorithm}
\usepackage[noend]{algorithmic}
\usepackage{amsmath}
\usepackage{amssymb}
\usepackage{mathtools}
\usepackage{amsthm}

\theoremstyle{plain}
\newtheorem{theorem}{Theorem}[section]
\newtheorem{proposition}[theorem]{Proposition}
\newtheorem{lemma}[theorem]{Lemma}

\newtheorem{corollary}[theorem]{Corollary}
\theoremstyle{definition}
\newtheorem{definition}[theorem]{Definition}
\newtheorem{assumption}[theorem]{Assumption}

\theoremstyle{remark}

\newcommand{\argmin}{\mathop{\rm argmin}}

\newcommand{\R}{\mathbb{R}}

\newcommand{\cone}{\text{cone}}
\renewcommand{\parallel}{\kern 0.56em/\kern -0.8em /\kern 0.56em}
\newcommand{\notparallel}{\kern 0.56em/\kern -0.8em /\kern -0.7em \smallsetminus\kern 0.09em}

\begin{document}



\twocolumn[

\aistatstitle{Preventing Model Extraction in Contextual Linear Optimization}

\aistatsauthor{Adam N. Elmachtoub \And Tianyi Lin \And Wenxuan Liu}

\aistatsaddress{Columbia University} ]

\begin{abstract}
Consider a contextual linear optimization problem, where a predict-then-optimize (PTO) system returns optimal decisions without revealing the cost vectors used in its downstream linear program. Our focus is on how to prevent attackers, who can submit context queries, from extracting the underlying prediction model. Although the cost-to-decision map is many-to-one, observations across contexts can reveal the underlying prediction model. For linear models, we provide sufficient conditions where the attacker can learn the model up to positive scaling by only observing the decision. These conditions motivate an adaptive attack that alternates inverse optimization with boundary sampling, i.e., using context queries leading to  decision boundaries of the current attacker model. To prevent extraction, we propose a mechanism that perturbs decisions only when the predicted cost vector is near a decision boundary. We establish a regret bound in terms of the perturbation radius. For the attackers that fit the defended responses as optimal decisions under the evaluation distribution, we also prove a lower bound on expected extraction error.  Experiments on synthetic instances and real data quantify the security-performance trade-off of the defense mechanism and illustrate that our defense can increase the extraction error. 
\end{abstract}

\section{Introduction}
Many decision-making problems can be formulated as optimization problems whose parameters are unknown and must be inferred from contexts. For example, a recommendation system uses user features to predict click probabilities before optimizing which items to display. In portfolio optimization, asset returns must be estimated before making allocation decisions. This class of problems is referred to as \textit{contextual optimization}~\citep{Sadana-2025-Survey}. A widely adopted approach to contextual optimization is the \textit{predict-then-optimize} (PTO) framework~\citep{Elmachtoub-2022-Smart}. In the linear optimization setting studied here, PTO proceeds in two stages: (i) a machine learning (ML) model predicts the unknown objective cost vector from a given context and (ii) the predicted cost vector is used to solve the downstream linear optimization problem and produce a decision. Due to its modularity and empirical effectiveness, PTO has been successfully applied across a broad range of application domains, including recommendation systems, e-commerce, logistics, and resource allocation.

This setting raises a model-security question: can an attacker recover the hidden prediction model by interacting only with the optimization system's decision outputs? Such a threat is known as a \textit{model extraction} attack: the attacker repeatedly queries the deployed system with contexts (perhaps through an API) and uses the returned decisions to build a substitute prediction model. Successful extraction can allow the attacker to replicate the service while bypassing access costs or usage restrictions, and moreover, can  provide a stepping stone for user privacy or inference attacks~\citep{Narayanan-2008-Robust, Papernot-2017-Practical, Shokri-2017-Membership, Salem-2019-ML, Juuti-2019-Prada} and information leakage \citep{Tan-2016-Information, Zhang-2023-Ignorance}. At first glance, the optimization layer appears to make extraction difficult. The attacker does not observe predicted costs directly, and the map from costs to optimal decisions is many-to-one in linear optimization. This intuition is incomplete: the same polyhedral geometry that hides costs in the interior of normal cones creates \textit{boundary contexts} where small changes in predicted costs change the returned decision. This leads to two central questions: (i) when do decision-level observations identify the underlying prediction model, and (ii) how can we mitigate extraction while preserving decision quality?

Existing approaches do not answer these questions directly. Prior work on privacy in contextual decision-making has studied differential privacy (DP)~\citep{Lei-2024-Privacy}, which protects training data but does not directly defend the deployed predictive model against extraction. A second related line is inverse optimization~\citep{Chan-2025-Inverse}, which infers unknown parameters of an optimization problem from observed optimal decisions. While inverse optimization includes both fixed-parameter and
contextual formulations, our focus is on whether decision-level observations identify the deployed predictor. More importantly, because the decision map is many-to-one, decision consistency does not in general imply prediction recovery.

In this paper, we focus on model extraction in contextual linear optimization. First, we study vulnerability through a loss calibration condition that links global inverse feasibility consistency to prediction recovery. Motivated by this condition, we then develop an adaptive attack that combines inverse optimization with boundary sampling to reconstruct the hidden predictor using only decision-level observations. This insight motivates a defense that perturbs the returned decision only when the deployed predictor's cost vector is near a decision boundary. Our theory establishes a regret bound for linear predictors and a lower bound on expected prediction error under a response-matching condition. Our experiments on synthetic instances and real data examine the security-performance trade-off.

\textbf{Contributions.} We formalize model extraction in contextual linear optimization under the predict-then-optimize paradigm and study when the decision-level observations identify the predictor and how to mitigate extraction while limiting the loss in decision quality. Our contributions are summarized as follows:
\begin{itemize}
\item We identify loss calibration as a vulnerability criterion: if it holds, global inverse feasibility consistency is sufficient for prediction recovery up to positive scaling. For linear predictors and unrestricted contextual queries, we use boundary contexts to establish this identification result under sufficient geometric and regularity conditions. This motivates an adaptive attack that alternates inverse optimization with boundary sampling, which targets contexts near decision boundaries of the current attacker model.
\item We propose a decision perturbation defense that randomizes only near decision boundaries and prove a regret upper bound that is linear in the boundary radius. For attackers that fit the defended
responses as optimal decisions under the evaluation distribution, we also prove a lower bound on expected prediction error. This lower bound is nondecreasing in the boundary radius.
\item We evaluate the attack and defense on synthetic instances and a short video recommendation task. In the tested settings, boundary sampling achieves lower prediction error than the non-adaptive inverse-optimization baseline, while the proposed defense can increase the prediction error of the evaluated attacks at modest regret.
\end{itemize}
\textbf{Related work.} Privacy in contextual optimization has been studied through differential privacy, including personalized pricing and revenue management~\citep{Chen-2022-Privacy, Chen-2023-Differential,Lei-2024-Privacy,Mochaourab-2021-Robust, Sun-2025-DP, Chowdhury-2023-Differentially, Fu-2026-Privacy, Selvi-2026-Differential}. These works are complementary to ours: DP protects data privacy, whereas we study model extraction attacks that target the deployed predictor itself. In model extraction attacks, the goal is to recover a substitute model via query access to a target system~\citep{Wang-2018-Stealing, Reith-2019-Efficiently, Orekondy-2019-Knockoff, Milli-2019-Model, Chandrasekaran-2020-Exploring, Pal-2020-Activethief, Truong-2021-Data, Huang-2024-Enhancing, Ferry-2024-Trained, Khouna-2025-Counterfactuals, Gorge-2026-Training}. Our work is also related to adaptive query attacks~\citep{Tramer-2016-Stealing, Zhao-2025-Systematic}. In our setting, the attacker observes downstream decisions but not predicted cost vectors. Since the prediction-to-decision map
is generally many-to-one, we examine how the geometry of the downstream feasible set affects predictor recovery.

Another related line concerns inverse optimization, which infers unknown problem parameters from observed decisions~\citep{Chan-2014-Generalized, Aswani-2018-Inverse, Babier-2021-Ensemble, Sun-2023-Maximum, Besbes-2025-Contextual, Chan-2025-Inverse, Patel-2026-Tight}. We use inverse optimization to study when decision consistency identifies the underlying predictor.

\textbf{Notation.} We write $x \parallel y$ for parallel vectors and $\angle(x,y)$ for the angle between two nonzero vectors. The polyhedral cone generated by $a_1,\ldots,a_k$ is denoted by $\operatorname{polycone}\{a_1,\ldots,a_k\}$. For a polytope $S \subseteq \R^n$, we let $V$ denote its vertex set. For a vertex $w \in V$, we denote $\cone(w)$ as its normal cone: the set of cost vectors $c \in \R^n$ satisfying $c^\top(u-w) \geq 0$ for all $u \in S$. We also denote by $N_V(w) = \{v\in V:\dim(\cone(v) \cap \cone(w)) = n-1\}$ the set of vertices whose normal cones share a facet with $\cone(w)$. For $c \neq 0$, its angular distance to $\cone(w)$ is $\angle(c,\cone(w)) = \inf_{u \in \cone(w) \setminus \{0\}} \angle(c,u)$.

\section{Model Formulation}
We study model extraction in contextual linear optimization. This setting involves three roles: a decision-maker, an attacker, and some users. For each user, the decision-maker maps contexts to an action through a predict-then-optimize (PTO) pipeline. Given a cost vector $c\in\R^n$, the downstream decision is obtained by solving a linear program in the form of
\begin{equation}\label{eq:contextOpt}
z^\star(c)=\min_{w\in S} c^\top w,
\end{equation}
where $S=\{w \in \R^n: Aw \leq b\}$ is a compact, full-dimensional polytope. We assume that each inequality in $Aw \leq b$ defines a distinct facet of $S$, and denote the nonzero rows of $A$ by
$a_1^\top,\ldots,a_m^\top$. We denote the optimal solution set by $W^\star(c)=\argmin_{w \in S} c^\top w$ and let $w^\star(c) \in W^\star(c) \cap V$ be an optimal vertex selected by a fixed tie-breaking rule. Compactness ensures existence of an optimizer, while full dimensionality allows us to study cost directions in the ambient space\footnote{For a lower-dimensional feasible set, an affine-hull reduction preserves the optimization problem~\citep{Rockafellar-1970-Convex}, but recovery concerns only the predictor projected onto the direction space of that affine hull.}.

In the PTO pipeline, the user's true cost vector is not observed. The decision-maker instead learns a prediction model $\hat c=p(x)$ from contexts $x\in\R^d$ and exposes the induced decision rule $\tilde w(x)=w^\star(p(x))\in W^\star(p(x))$ (perhaps through an API)\footnote{Our conditions are set-valued: an attacker is consistent with the response at $x$ if $w^\star(P^{DM}x) \in W^\star(P^a x)$. This convention avoids making the analysis depend on arbitrary tie-breaking among multiple optimal vertices.}.  A user submits a context $x$ and receives the decision $\tilde w(x)$. We focus on the linear prediction model\footnote{Appendix~\ref{sec:app:exp} also evaluates the attack empirically under nonlinear deployed predictors and misspecified linear attacker models.} in the form of $p(x)=P^{DM}x$, where $P^{DM} \in \R^{n\times d}$ is the decision-maker's prediction matrix. The attacker knows the linear structure but not $P^{DM}$, and learns a model $p^a(x)=P^a x$ with $P^a\in\R^{n\times d}$. We consider $d > n \geq 2$, assume $\operatorname{rank}(P^{DM})=n$, and exclude the zero attacker matrix $P^a=0$. 

\textbf{Attackers.} A key step is to define the attacker's prediction loss. Since positive scaling of the cost vector does not change the optimal solution set of a linear program, the attacker succeeds whenever it recovers a prediction model $P^a=\gamma P^{DM}$ for some $\gamma>0$. We then measure prediction error by angular distance. Since $d>n$, prediction matrices have nontrivial kernels. We extend the angular loss by setting $\angle(0,0)=0$ and $\angle(u,v)=\frac{\pi}{2}$ when exactly one of $u$ and $v$ is zero.
\begin{definition}[Prediction Loss]\label{def:prediction}
Given a context $x$ and prediction matrices $P^a$ and $P^{DM}$, the prediction loss is $L_{\textnormal{pred}}(P^a,P^{DM},x)=\angle(P^a x,P^{DM}x)$. For $D=\{(x_i,w_i)\}_{i=1}^N$, the empirical prediction loss is $L_{\textnormal{pred}}(P^a,P^{DM};D)=\tfrac{1}{N} \sum_{i=1}^N \angle(P^a x_i,P^{DM}x_i)$. 
\end{definition}
We assume that the attacker can make $N$ queries $x_i \in \R^d$, each chosen using the preceding query-response pairs. Without defense, query $x_i$ returns $w_i=w^\star(p(x_i))$. After observing $D^a=\{(x_i,w_i)\}_{i=1}^N$, the attacker returns a prediction matrix $P^a = \mathcal{A}(D^a)$, where $\mathcal{A}$ is his algorithm. Fixing a test distribution $Q$, we evaluate $\mathbb{E}[L_\textnormal{pred}(P^a,P^{DM},x)]$, where the expectation is taken over an independent evaluation context $x \sim Q$ and any attack or defense randomness. The attacker aims to reduce this prediction loss using only information available through the query interface. The following proposition shows that traditional model extraction on classification model fits into the predict-then-optimize model.
\begin{proposition}\label{prop:generalize-classification}
For the simplex $S = \{w \in \R_+^n: \sum_{j=1}^n w_j=1\}$, we define $p(x)$ as the negative class utility vector of an $n$-class classifier. With matched tie-breaking rules, returning the optimal vertex $w^\star(p(x))=e_i$ is equivalent to returning label $i$ for a $n$-classification problem with prediction model $p(x)$. 
\end{proposition}

\textbf{Defense.} To defend against extraction, the decision-maker may strategically modify the query interface to restrict the information revealed to adversarial users. Instead of returning the exact optimizer $w^\star(p(x))$, the decision-maker (API) outputs a perturbed decision $\tilde{w}(x)$, so that an attacker querying $x_i$ observes $w_i=\tilde{w}(x_i)$. The decision-maker aims to prevent reconstruction of the prediction model while maintaining near-optimal decision quality.

This interaction admits a Stackelberg interpretation: the decision-maker commits to a response mechanism before the attacker chooses its queries and learning procedure. For a specified attack,
the security-performance trade-off is measured by
\begin{equation*}
\begin{aligned}
\textbf{Security: } & \mathbb{E}[L_\textnormal{pred}(P^a,P^{DM},x)],\\
\textbf{Performance: } & \mathbb{E}[p(x)^\top(\tilde{w}(x)-w^\star(p(x)))].
\end{aligned}
\end{equation*}
The expectations are taken over an independent $x \sim Q$ and any attack or defense randomness. The first criterion favors larger attacker prediction error, whereas the second favors smaller regret under
the deployed predicted costs. We consider the expected regret when $\mathbb{E}_{x \sim Q}\|p(x)\|_2<+\infty$.

\section{Model Extraction}\label{sec:attacker}
We characterize how decision-level query responses can reveal the underlying prediction model. Throughout this section, we assume that no defensive mechanism is deployed, so the decision-maker returns the true optimal decision $\tilde w(x)=w^\star(P^{DM}x)$. 

\subsection{Vulnerability identification} \label{subsec:vulnerability}
The attacker first assesses whether the decision-maker is vulnerable to model extraction, where vulnerability means that the prediction model can be recovered using decision-level queries. Since $P^{DM}x$ is never observed, the attacker cannot directly evaluate $L_{\textnormal{pred}}(\cdot,P^{DM},x)$. Instead, the attacker optimizes a surrogate loss that depends on $P^{DM}$ only through the observed decision $w^\star(P^{DM}x)$. A class of surrogate loss functions comes from inverse optimization~\citep{Chan-2025-Inverse}. Given a query-response dataset $D=\{(x_i,w_i)\}_{i=1}^N$, we define the inverse feasible set as follows:
\begin{equation*}
\mathcal{P}_\textnormal{inv}(D)=\{P \neq 0^{n \times d}: w_i\in W^\star(Px_i), \ \forall (x_i,w_i)\in D\}.
\end{equation*}
We allow $Px_i=0$ at individual contexts but exclude the zero prediction matrix. This is the set of prediction matrices whose induced costs make every observed decision optimal.
\begin{definition}
A nonnegative loss $L(\cdot,P^{DM};D)$ defined on nonzero matrices is an inverse feasibility loss for $D$ if $L(P,P^{DM};D)=0 \Longleftrightarrow P \in \mathcal{P}_\textnormal{inv}(D)$. We also write $L(P,P^{DM};x) = L(P,P^{DM};\{(x,w^\star(P^{DM}x))\})$ for a single context $x \in \R^d$. 
\end{definition}
Achieving zero inverse feasibility loss enforces \emph{decision consistency} on the observed queries. However, decision consistency alone does not in general imply prediction recovery since different cost vectors can induce the same optimal decision. We thus need to identify when inverse feasibility losses are appropriate surrogates for the prediction loss.
\begin{definition}\label{def:vulnerable-condition}
An inverse feasibility loss calibrates the prediction loss if, for any attacker model $P^a$, we have that $L(P^a,P^{DM};x)=0$ for all $x\not\in \ker(P^{DM})$ if and only if $L_{\textnormal{pred}}(P^a,P^{DM};x)=0$ for all $x\not\in \ker(P^{DM})$. Equivalently, global decision consistency implies recovery of the predicted cost directions, up to the positive scaling ambiguity of linear objectives.
\end{definition}
Loss calibration is an exact global identification property: zero inverse feasibility loss on all nonkernel contexts identifies the predictor up to positive scaling. It does not itself bound prediction error from a small loss on a finite dataset. The following lemma shows that for a fixed downstream problem, calibration holds for either all inverse feasibility losses or none.
\begin{lemma}\label{lem:invariant-calibration}
The loss calibration property holds for either all inverse feasibility losses or none. If $S$ is an $n$-dimensional hypercube, the loss calibration property holds for none. 
\end{lemma}
We provide a sufficient condition under which the loss calibration property holds in the following assumption. 
\begin{assumption}\label{assumption:main}
There exist $n+1$ facet normals $\{a_{k_i}\}_{i=1}^{n+1}$ so that $a_{k_1},\ldots,a_{k_n}$ are linearly independent and $a_{k_{n+1}}=\sum_{i=1}^n \beta_i a_{k_i}$ where $\beta_i \neq 0$ for all $i$.
\end{assumption}
\begin{theorem}\label{thm:loss-cali-cond}
Suppose Assumption~\ref{assumption:main} holds. Then, if a nonzero attacker model $P^a$ satisfies $w^\star(P^{DM}x) \in W^\star(P^a x)$ for all $x \notin \ker(P^{DM})$, we have $P^a=\gamma P^{DM}$ for some $\gamma>0$. Consequently, every inverse feasibility loss calibrates the prediction loss.
\end{theorem}
The geometry behind Theorem~\ref{thm:loss-cali-cond} also motivates the use of \textit{boundary contexts}. Indeed, we define $r_i=-\frac{a_i}{\|a_i\|}$. A boundary context associated with facet $i$ satisfies $P^a\bar{x}=\alpha r_i$ for some $\alpha>0$. At this cost, every vertex of that facet is optimal.
For an attacker model, nearby contexts can enter each incident normal cone. Global decision consistency constrains the true cost direction at these boundary contexts.
\begin{proposition}\label{prop:edge-sample}
Suppose that the rank of $P^a$ is $n$ and $w^\star(P^{DM}x) \in W^\star(P^a x)$ for all $x \notin \ker(P^{DM})$. Then, $L_\textnormal{pred}(P^a,P^{DM},\bar{x})=0$ at every boundary context $\bar{x}$. 
\end{proposition}
Proposition~\ref{prop:edge-sample} shows that global decision consistency fixes the cost direction at boundary contexts. To establish Theorem~\ref{thm:loss-cali-cond}, global consistency first gives $P^a=BP^{DM}$, and cone preservation implies $Ba_i=\lambda_i a_i$ with $\lambda_i\geq0$. Applying $B$ to the relation in Assumption~\ref{assumption:main} gives $\beta_i(\lambda_{k_i}-\lambda_{k_{n+1}})=0$ for every $i$. Since all $\beta_i$ are nonzero, these eigenvalues coincide, so $B$ is a scalar multiple of the identity. The scalar is positive because $P^a \neq 0$.

Assumption~\ref{assumption:main} holds for various application problems with $S = \{w: 0 \leq w \leq 1, a^\top w\le b\}$ when $a_i>0$ for all $i$ and $0<b<\sum_i a_i$. The lower-coordinate facets have normals $-e_1,\ldots,-e_n$ and the budget facet has normal $a$. Since $a=\sum_i (-a_i)(-e_i)$, this assumption holds. By contrast, the hypercube has the normals $\{\pm e_i\}_{i=1}^n$ and does not satisfy Assumption~\ref{assumption:main}. Although failing a sufficient condition does not prove that calibration fails, Lemma~\ref{lem:invariant-calibration} shows that calibration fails for this set. For comparison, on the Euclidean unit ball, the optimal decision is $w^\star(c)=-\frac{c}{\|c\|}$ for $c \neq 0$, so it reveals the cost direction. This implies that vulnerability is not restricted to the polyhedral setting.
\begin{algorithm}[!t] \small
\caption{Attacking scheme}\label{algo:attack}
\begin{algorithmic}[1]
\STATE \textbf{Input:} $t=0$, $P^0 \neq 0^{n \times d}$, $\Delta>0$ and $L(P,P^{DM};D)$.
\STATE \textbf{Initialize:} $P^0 \leftarrow \frac{P^0}{\|P^0\|_F}$ and $D^0 \leftarrow \textnormal{Sampling}(P^0)$.
\WHILE{there exist $(x_i,w_i) \in D^t$ and a vertex $w_k$ such that $(P^t x_i)^\top(w_i-w_k) > \Delta$}
\STATE $P^{t+1} \in \argmin_{P \neq 0} L(P,P^{DM};D^t)$. 
\STATE $P^{t+1} \leftarrow \frac{P^{t+1}}{\|P^{t+1}\|_F}$.
\STATE $D^{t+1} \leftarrow D^t\cup \textnormal{Sampling}(P^{t+1})$. 
\STATE $t \leftarrow t+1$.
\ENDWHILE
\STATE \textbf{Return:} $P^t$.
\end{algorithmic}
\end{algorithm}

\subsection{Algorithmic framework}
Under Assumption~\ref{assumption:main}, global decision consistency implies prediction recovery. This motivates the adaptive attack in Algorithm~\ref{algo:attack} that alternates inverse optimization with sampling. The algorithm stops when every observed decision is $\Delta$-optimal under the current attacker model $P^t$. The following proposition shows that this procedure stops after finite steps.
\begin{proposition}\label{prop:adv-algo-stop}
Suppose $\|P^0\|_F=1$, each inverse-optimization update achieves zero inverse feasibility loss, and every sampling call returns finitely many queries. Assume $\|x_i\| \leq R$ for every query in the run, for some finite $R>0$. For $\Delta>0$, Algorithm~\ref{algo:attack} terminates after at most $(1+\frac{2RL_e}{\Delta})^{nd}$ inverse optimization updates, where $L_e = \max_{u,v \in V} \|u-v\|$.
\end{proposition}
The tolerance $\Delta$ controls empirical suboptimality after Frobenius normalization. The bound counts inverse optimization updates and our result guarantees finite termination. A smaller $\Delta$ makes the stopping criterion stricter and may require more updates.

\textbf{Inverse optimization.} The inverse optimization step minimizes an inverse feasibility loss on the current dataset $D^t$. Any inverse feasibility loss can be used in this framework and we use the suboptimality loss $L_\textnormal{SO}(P,P^{DM},x_i)=(Px_i)^\top(w_i-w^\star(Px_i))$ where $w_i=w^\star(P^{\mathrm{DM}}x_i)$ is the observed decision. We define $L_\textnormal{SO}(P,P^{DM};D) = \sum_{(x_i,w_i) \in D} L_\textnormal{SO}(P,P^{DM},x_i)$. The empirical loss is zero exactly when $w_i \in W^\star(Px_i)$ for every observation.~\citet{Babier-2021-Ensemble} have shown that minimizing the empirical suboptimality loss admits the convex reformulation in the form of
\begin{equation} \label{eq:subopt-reform}
\begin{array}{rcl}
& \min\limits_{P,\lambda_i} & \sum_{i=1}^N |(Px_i)^\top w_i-b^\top\lambda_i|, \\ 
& \textnormal{s.t. } & A^\top\lambda_i=Px_i, \lambda_i \leq 0 \textnormal{ for all } i.
\end{array}
\end{equation}
The above homogeneous formulation admits $P=0$. In the experiments, we solve two convex optimization problems with the respective constraints $\sum_{i=1}^n \sum_{j=1}^d P_{ij} \geq 1$ or $\sum_{i=1}^n \sum_{j=1}^d P_{ij}\leq -1$ before normalizing the fitted matrix in Frobenius norm. These constraints exclude zero prediction matrices but do not cover matrices whose entries sum to zero. Appendix~\ref{sec:app-io-loss-opt} discusses this restriction and alternative inverse feasibility losses.

\textbf{Sampling.} Decision consistency on a finite dataset does not necessarily imply small prediction loss, even under loss calibration, since many cost vectors can induce the same downstream optimizer. The sampling step therefore focuses on contexts where this ambiguity is easiest to expose.

\usetikzlibrary{calc}
\usetikzlibrary{shapes.geometric}

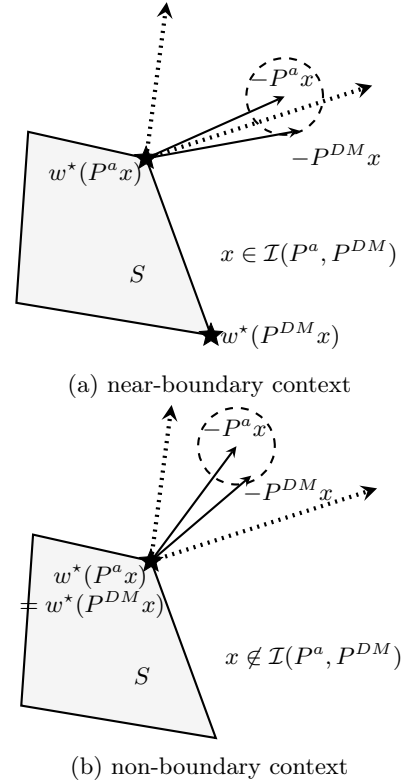
\begin{figure}[!t]
\centering

% =========================
% (a) near-boundary context
% =========================
\begin{subfigure}[t]{0.35\textwidth}
\centering
\begin{tikzpicture}[scale=0.78, >=stealth]

\tikzset{
    poly/.style={draw=black, thick},
    normal/.style={->, dashed, thick},
    cost/.style={->, thick},
    ball/.style={dashed, thick},
    every node/.style={font=\small},
    every node/.style={font=\footnotesize}
}

% ---------- key points ----------
\coordinate (v) at (0,0);
\coordinate (A) at (-2.0,0.45);
\coordinate (B) at (-2.2,-2.45);
\coordinate (C) at (1.10,-3.00);

% normal directions
\coordinate (n1) at (0.35,2.65);
\coordinate (n2) at (3.835,1.235);

% cost vectors
\coordinate (Pa)  at (2.35,1.05);
\coordinate (Pdm) at (2.60,0.47);

% ---------- polytope S ----------
\fill[gray!8] (A) -- (v) -- (C) -- (B) -- cycle;
\draw[poly] (A) -- (v) -- (C) -- (B) -- cycle;

\node at (-0.15,-1.95) {$S$};

% ---------- vertices ----------
\node[star, star points=5, star point ratio=2.25,
      fill=black, draw=black, minimum size=9pt,
      inner sep=0pt] at (v) {};

\node[star, star points=5, star point ratio=2.25,
      fill=black, draw=black, minimum size=9pt,
      inner sep=0pt] at (C) {};

% ---------- labels for optimal vertices ----------
\node[anchor=north east] at (0.1,0.1) {$w^\star(P^a x)$};
\node[anchor=west] at (C) {$w^\star(P^{DM} x)$};

% ---------- normal vectors ----------
\draw[normal, dotted, ultra thick] (v) -- (n1);
\draw[normal, dotted, ultra thick] (v) -- (n2);

% ---------- cost vectors ----------
\draw[cost] (v) -- (Pa)
    node[pos=0.98, above] {$-P^a x$};

\draw[cost] (v) -- (Pdm)
    node[pos=0.88, below right] {$-P^{DM} x$};

% ---------- dashed circle ----------
\draw[ball] (Pa) circle (0.65);

% --------- x in I(P,PDM) ------

\node[anchor=south west] at (1.1,-2) {$x\in\mathcal{I}(P^a,P^{DM})$};

\end{tikzpicture}
\caption{near-boundary context}
\label{subfig:near-boundary-context}
\end{subfigure}
\hfill
% =========================
% (b) non-boundary context
% =========================
\begin{subfigure}[t]{0.35\textwidth}
\centering
\begin{tikzpicture}[scale=0.78, >=stealth]

\tikzset{
    poly/.style={draw=black, thick},
    normal/.style={->, dashed, thick},
    cost/.style={->, thick},
    ball/.style={dashed, thick},
    every node/.style={font=\small},
    every node/.style={font=\footnotesize}
}

% ---------- key points ----------
\coordinate (v) at (0,0);
\coordinate (A) at (-2.0,0.45);
\coordinate (B) at (-2.2,-2.45);
\coordinate (C) at (1.10,-3.00);

% normal directions
\coordinate (n1) at (0.35,2.65);
\coordinate (n2) at (3.835,1.235);

% cost vectors: both inside the normal cone
\coordinate (Pa)  at (1.45,1.95);
\coordinate (Pdm) at (1.70,1.45);

% ---------- polytope S ----------
\fill[gray!8] (A) -- (v) -- (C) -- (B) -- cycle;
\draw[poly] (A) -- (v) -- (C) -- (B) -- cycle;

\node at (-0.15,-1.95) {$S$};

% ---------- vertices ----------
\node[star, star points=5, star point ratio=2.25,
      fill=black, draw=black, minimum size=9pt,
      inner sep=0pt] at (v) {};

% ---------- labels for optimal vertices ----------
\node[anchor=north east] at (0.1,0.1) {$w^\star(P^a x)$};
\node[anchor=north east] at (0.4,-0.35) {$=w^\star(P^{DM} x)$};

% ---------- normal vectors ----------
\draw[normal, dotted, ultra thick] (v) -- (n1);
\draw[normal, dotted, ultra thick] (v) -- (n2);

% ---------- cost vectors ----------
\draw[cost] (v) -- (Pa)
    node[pos=0.98, above] {$-P^a x$};

\draw[cost] (v) -- (Pdm)
    node[pos=0.82, right] {$-P^{DM} x$};

% ---------- dashed circle ----------
\draw[ball] (Pa) circle (0.65);

% --------- x in I(P,PDM) ------

\node[anchor=south west] at (1.1,-2) {$x\not\in\mathcal{I}(P^a,P^{DM})$};

\end{tikzpicture}
\caption{non-boundary context}
\label{subfig:interior-context}
\end{subfigure}

\caption{Illustration of prediction misalignment. Arrows show negative cost vectors. The two models return different decisions in (\subref{subfig:near-boundary-context}) and the same decision in (\subref{subfig:interior-context}). }
\label{fig:context} \vspace{-1.5em}
\end{figure}

Our key idea is to query boundary and near-boundary contexts. Using the directions $r_i$ defined in Section~\ref{subsec:vulnerability}, an $\epsilon$-near-boundary context satisfies $P^a x \neq 0$ and
$\angle(P^a x,r_i) < \epsilon$ for some $i$. Such queries can expose prediction misalignment even when interior contexts return the same decision (see Figure~\ref{fig:context}). Formally, we define $I(P^a,P^{DM})= \{x: L_\textnormal{SO}(P^a,P^{DM},x)>0\}$. A context in this set reveals a decision inconsistency. The following proposition summarizes the observation. 
\begin{proposition}\label{prop:info-near-boundary}
Suppose that Assumption~\ref{assumption:main} holds, the rank of $P^a$ is $n$ and $L_\textnormal{pred}(P^a,P^{DM},x)>0$ for some $x$. Then, for every $\epsilon>0$, there is an $\epsilon$-near-boundary context $\tilde{x} \in I(P^a,P^{DM})$.
\end{proposition}
Proposition~\ref{prop:info-near-boundary} shows that, whenever the attacker model $P^a$ is misaligned with $P^{DM}$, some near-boundary query can expose this misalignment through nonzero inverse feasibility loss. Boundary sampling is designed to choose informative queries. Given $P^a$, the attacker queries contexts whose predicted costs point along the inward facet directions $r_i$, and queries nearby contexts obtained by moving slightly from these boundary directions into adjacent normal cones. We use a parameter $M$ to control proximity to these directions, where larger $M$ places the sampled cost directions closer to the corresponding facet rays. Appendix~\ref{sec:app:sampling} specifies the boundary sampling construction.

We now state the convergence guarantee for Algorithm~\ref{algo:attack} with boundary sampling. 
\begin{theorem}[Informal]\label{thm:sample-global-error-informal}
Suppose Assumption~\ref{assumption:main} holds. Run Algorithm~\ref{algo:attack} with a full-rank initialization and boundary sampling as specified in Appendix~\ref{sec:app:sampling}. Assume that each inverse optimization update returns a full-rank matrix with zero inverse feasibility loss on the accumulated dataset, and that the samples satisfy the uniform boundedness and conditioning requirements of Theorem~\ref{thm:sample-global-error} across all runs. Choose $M=M(\Delta)$ with $M(\Delta) \to +\infty$ as $\Delta \downarrow 0$, and suppose that $(M,\Delta)$ satisfies the stopping condition in Theorem~\ref{thm:sample-global-error}. Each run terminates, and its output satisfies $L_{\textnormal{pred}}(P^a_{M(\Delta),\Delta},P^{DM},x) \to 0$ for every $x \notin \ker(P^{DM})$.
\end{theorem}
Theorem~\ref{thm:sample-global-error-informal} establishes asymptotic prediction recovery under
uniform sampling regularity. This regularity prevents the boundary representatives from becoming asymptotically uninformative. The stopping condition couples $M$ and $\Delta$ so that the algorithm cannot stop while a prescribed near-boundary response disagrees with its unique predicted optimizer. Appendix~\ref{sec:app:sampling} gives the construction and the formal conditions.

\section{Preventing Model Extraction}
We introduce a \emph{prediction perturbation} baseline that adds Gaussian noise to the predicted cost before optimization (see Algorithm~\ref{algo:PPDM-strategy}). This scheme perturbs costs without using their distance to a decision boundary. Independent resampling allows an attacker to estimate the response distribution by repeating a query.
\begin{algorithm}[!t] \small
\caption{Prediction perturbation scheme} \label{algo:PPDM-strategy}
\begin{algorithmic}[1]
\STATE \textbf{Input:} $x$, $P^{DM}$ and $\sigma$.
\STATE Sample $\eta \sim \mathcal{N}(0,\sigma^2 I_n)$.
\STATE \textbf{Return:} $\tilde{w}(x)=w^\star(P^{DM}x+\eta)$.
\end{algorithmic}
\end{algorithm}

We then propose a \emph{decision perturbation} defense. Since near-boundary queries can expose prediction misalignment, we modify responses in these regions by selecting neighboring decisions whose normal cones are close to the predicted cost. Algorithm~\ref{algo:DM-strategy} first computes the true optimizer and then checks whether the predicted cost $P^{DM}x$ lies close to the normal cone of any neighboring vertex. If no neighboring normal cone is within angular distance $\epsilon$, $\mathcal{W}=\{w_0\}$ and the decision-maker returns the true optimal decision. Otherwise, the algorithm returns a vertex from $\mathcal{W}$, which contains $w_0$ and all eligible neighboring vertices. For each fixed deployed model and boundary radius, the cache is initialized once and retained across queries. Each new candidate set is assigned a uniformly sampled member using fresh randomness. The cache prevents repeated queries from yielding fresh independent perturbations: once a candidate set $W$ has been assigned an output, the same output is returned whenever that set appears again. 
\begin{algorithm}[!t] \small
\caption{Decision perturbation scheme}\label{algo:DM-strategy}
\begin{algorithmic}[1]
\STATE \textbf{Input:} $x$, $P^{DM}$ and $\epsilon \geq 0$.
\STATE Compute $c=P^{DM}x$ and $w_0=w^\star(c)$. 
\IF{$c=0$ or $\epsilon=0$} 
\STATE \textbf{Return:} $w_0$. 
\ENDIF
\STATE \textbf{Initialize:} $\mathcal{W}=\{w_0\}$.
\FOR{$w\in N_V(w_0)$}
\IF{$\angle(c,\cone(w)) \leq \epsilon$} 
\STATE $\mathcal{W} \gets \mathcal{W} \cup \{w\}$. 
\ENDIF
\ENDFOR
\IF{$\mathcal{W}$ has been recorded} 
\STATE \textbf{Return:} $\textnormal{record}(\mathcal{W})$.
\ELSE
\STATE Sample $\tilde{w}$ uniformly from $\mathcal{W}$.
\STATE $\textnormal{record}(\mathcal{W}) \gets \tilde{w}$.
\STATE \textbf{Return:} $\tilde{w}$.
\ENDIF
\end{algorithmic}
\end{algorithm}

The parameter $\epsilon>0$ controls which neighboring decisions are eligible for perturbation. Increasing $\epsilon$ enlarges each candidate set. Below, we bound the regret incurred by the defense scheme in Algorithm~\ref{algo:DM-strategy} .
\begin{proposition} \label{prop:DM-regret}
Define the regret at context $x$ by $\textnormal{Regret}(w;x)=(P^{DM}x)^\top(w-w^\star(P^{DM}x))$.
Under Algorithm~\ref{algo:DM-strategy}, for every cache realization and every test distribution $Q$ with
$\mathbb{E}_{x \sim Q}[\|P^{DM}x\|]<+\infty$, we have
\begin{equation*}
\mathbb{E}_{x \sim Q}[\textnormal{Regret}(\tilde{w}(x);x)] \leq L_e\epsilon \mathbb{E}_{x \sim Q}[\|P^{DM}x\|],
\end{equation*}
where $L_e=\max_{u,v\in V}\|u-v\|$.
\end{proposition}
Proposition~\ref{prop:DM-regret} provides a regret upper bound that is linear in $\epsilon>0$. This bound also depends on the feasible set's diameter and the predicted cost's scale.

We quantify the prediction error incurred by attackers that fit the defended responses. For a context $x$, we write $c=P^{DM}x$ and $w_0=w^\star(c)$. For $c \neq 0$ and $\epsilon>0$, we define
\begin{equation*}
\mathcal{W}_\epsilon(x) = \{w_0\} \cup \{w\in N_V(w_0): \angle(c,\cone(w)) \leq \epsilon\}.
\end{equation*}
In this case, $\mathcal{W}_\epsilon(x)$ is the final candidate set in Algorithm~\ref{algo:DM-strategy} before cache lookup. When $c=0$ or $\epsilon=0$, we set $\mathcal{W}_\epsilon(x)=\{w_0\}$. For each fixed cache realization and boundary radius, we let $\mathcal{E}$ denote the set of predicted costs $c$ for which the algorithm returns a vertex different from $w^\star(c)$.
\begin{theorem} \label{thm:DM-def-BS}
For a context $x$ with $P^a x \neq 0$ and $P^{DM}x \neq 0$, we let $\phi(x)$ and $\psi(x)$ be the angular distances from these costs to the nearest normal cones neighboring $\cone(w^\star(P^a x))$ and $\cone(w^\star(P^{DM}x))$, respectively, with margin zero at ties. For every cache realization, if $\tilde{w}(x) \in W^\star(P^a x)$ and $P^{DM}x\in \mathcal{E}$, we have
\begin{equation*} 
L_\textnormal{pred}(P^a,P^{DM},x) \geq \psi(x)+\phi(x).
\end{equation*}
In particular, the loss is at least $\epsilon$ if also $\phi(x) \geq \epsilon$. The fitting condition holds in particular if
\begin{equation}\label{eq:fitting-condition}
(P^a x)^\top(\tilde{w}(x)-w^\star(P^a x)) < \|P^a x\| \ell_e \sin(\phi(x)), 
\end{equation}
where $\ell_e=\min_{u \neq v\in V} \|u-v\|$. Fix $Q$ independently of the cache. Suppose that, for almost every realization of the defense and attacker randomness, both costs are nonzero and $\tilde{w}(x) \in W^\star(P^a x)$ for $Q$-almost every $x$. Then, we have
\begin{equation*} 
\mathbb{E}[L_{\mathrm{pred}}(P^a,P^{\mathrm{DM}},x)] \geq B_\epsilon(Q) ,
\end{equation*}
where $B_\epsilon(Q) = \mathbb{E}_{x \sim Q}[\psi(x)(1-\frac{1}{|\mathcal{W}_\epsilon(x)|})]$. On the left-hand side, the expectation is taken over the defense and attacker randomness and an independent $x\sim Q$. For fixed $P^{DM}$ and $Q$, $B_\epsilon(Q)$ is nondecreasing in $\epsilon$.
\end{theorem}
Theorem~\ref{thm:DM-def-BS} formalizes the defense effect for response-fitting attackers: viewing a perturbed decision as optimal can lead to moving away from the true cost direction. Under the population fitting condition, it provides an expected prediction-error lower bound that is nondecreasing in $\epsilon$, while Proposition~\ref{prop:DM-regret} bounds the regret cost. Section~\ref{sec:exp} evaluates the effect of $\epsilon$ empirically.

\section{Experiments}\label{sec:exp}
We evaluate the proposed attack and defense methods on synthetic instances and a short-video recommendation task. We compare adaptive sampling attacks with inverse optimization baselines and examine the security-regret trade-off induced by the defensive algorithm. We report the angular prediction loss in Definition~\ref{def:prediction} and the decision-maker's regret in Proposition~\ref{prop:DM-regret}.

\begin{figure*}[!t]
\centering
\includegraphics[width=0.24\textwidth]{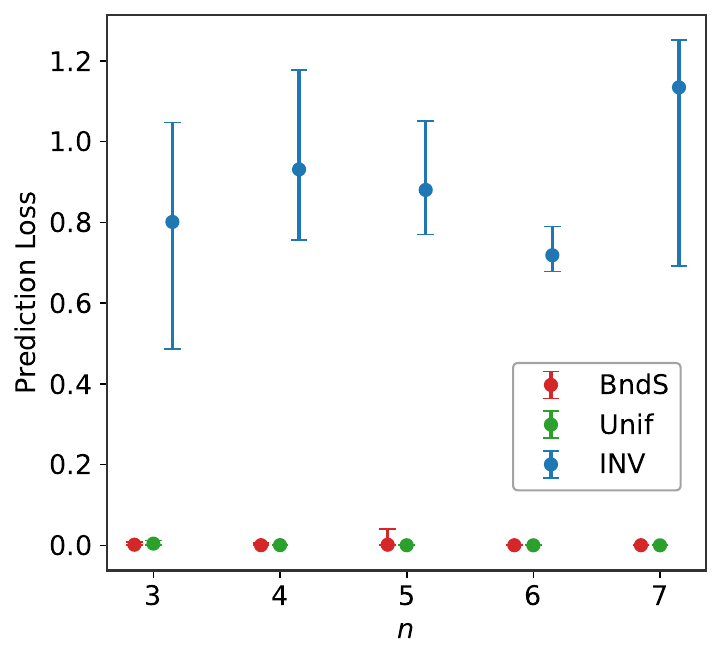}
\includegraphics[width=0.24\textwidth]{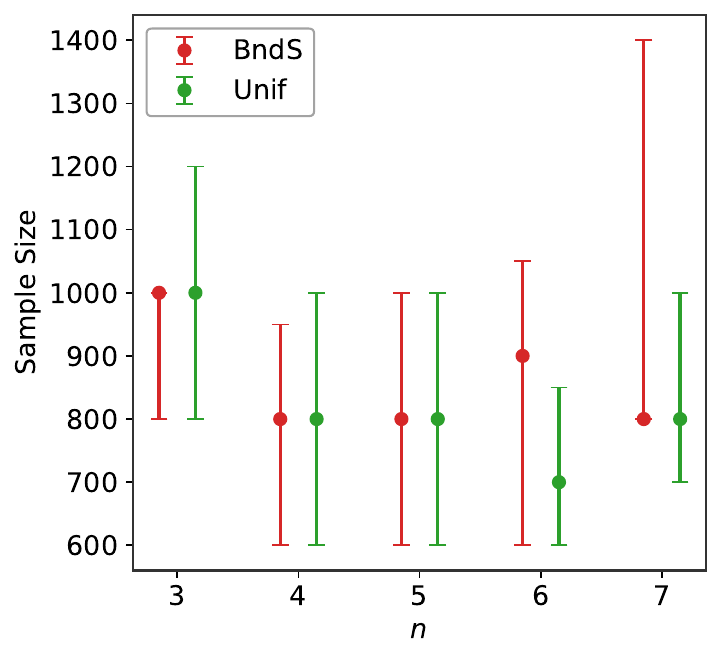}
\includegraphics[width=0.24\textwidth]{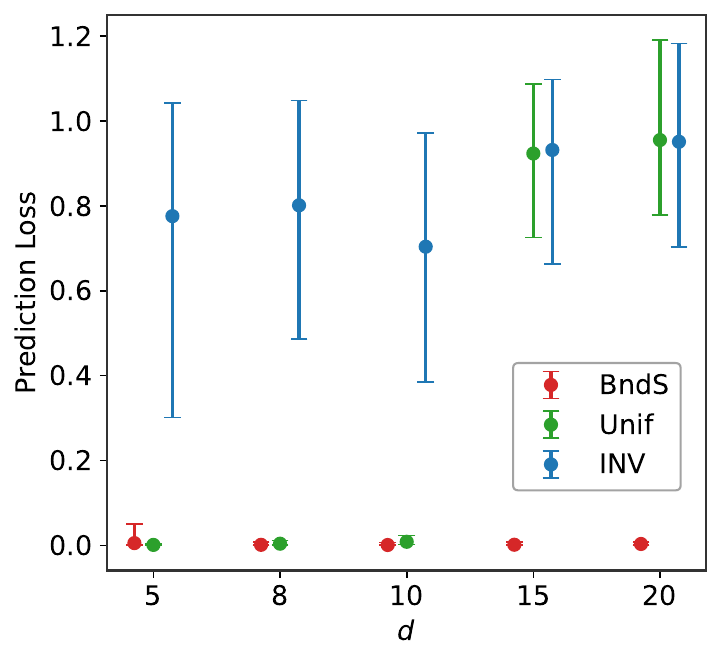}
\includegraphics[width=0.24\textwidth]{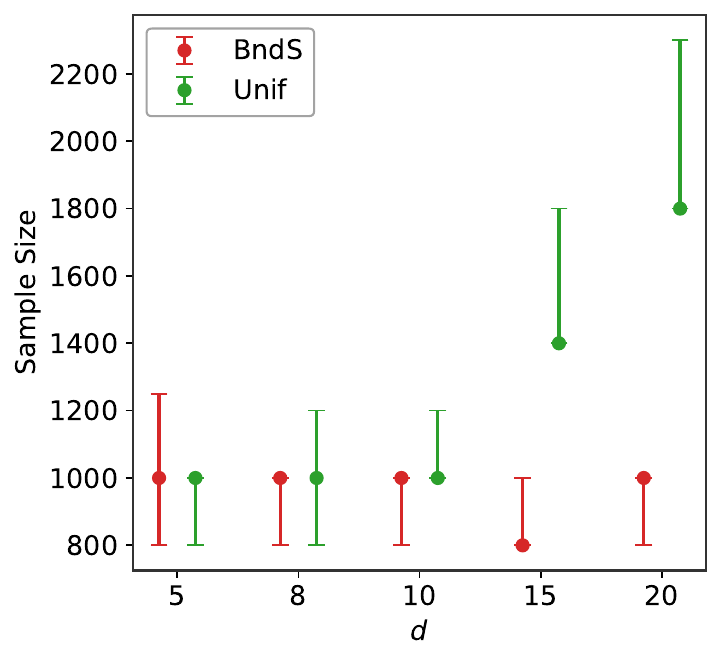}
\caption{Attack performance on synthetic data as the decision dimension $n$ (left) and feature dimension $d$ (right) vary. The first and third panels report prediction loss. The second and fourth report the number of context-decision observations required to attain loss below $0.001$. Prediction loss points and error bars show medians and $25^\textnormal{th}$-$75^\textnormal{th}$ percentiles over 100 repetitions per parameter setting.} \label{fig:1}
\end{figure*}

\textbf{Synthetic instances.} We compare three attacks: boundary sampling (BndS), uniform sampling (Unif), and an inverse optimization baseline method that uses a single nonadaptive fit (INV). Here, the feasible set is $S=\{w\in[0,1]^n:a^\top w\le b\}$, with $a_j=1+0.5(j-1)$ for $j=1,\ldots,n$ and $b=0.6\sum_{j=1}^n a_j$. The deployed model is fitted by linear regression to noisy synthetic data. Unif samples directions uniformly on the unit sphere and maps them to contexts using the current attacker model, so it remains adaptive. All three attacks use the suboptimality-loss fit in Eq.~\eqref{eq:subopt-reform}, and the INV comparisons use matched sample sizes. Figure~\ref{fig:1} reports prediction loss after three adaptive iterations for BndS and Unif, together with the INV comparison, and the sample sizes required to attain prediction loss below $0.001$ as the decision dimension $n$ and feature dimension $d$ vary. BndS achieves near-zero median prediction loss across the tested settings. Unif performs similarly at smaller feature dimensions, but has substantially higher loss at $d=15$ and $d=20$ after three iterations. BndS also uses fewer samples than Unif to reach the loss threshold at these larger feature dimensions. Across the tested decision dimensions, BndS has lower median prediction loss than INV in the displayed comparisons.

\begin{figure*}[!t]
\centering
\begin{subfigure}{0.32\textwidth}
\centering 
\includegraphics[width=\textwidth]{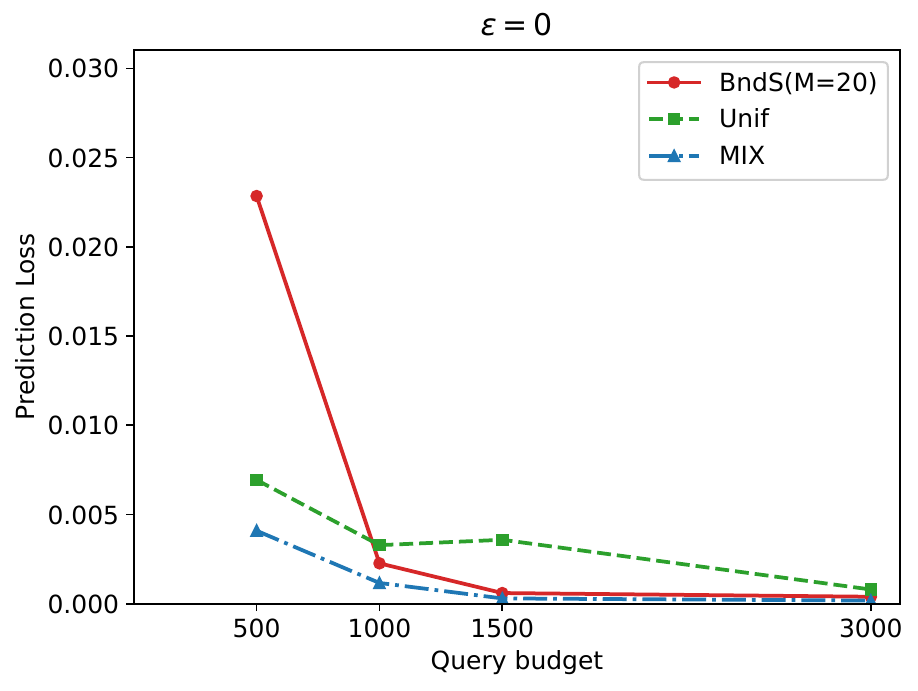}
\end{subfigure}
\begin{subfigure}{0.32\textwidth}
\centering
\includegraphics[width=\textwidth]{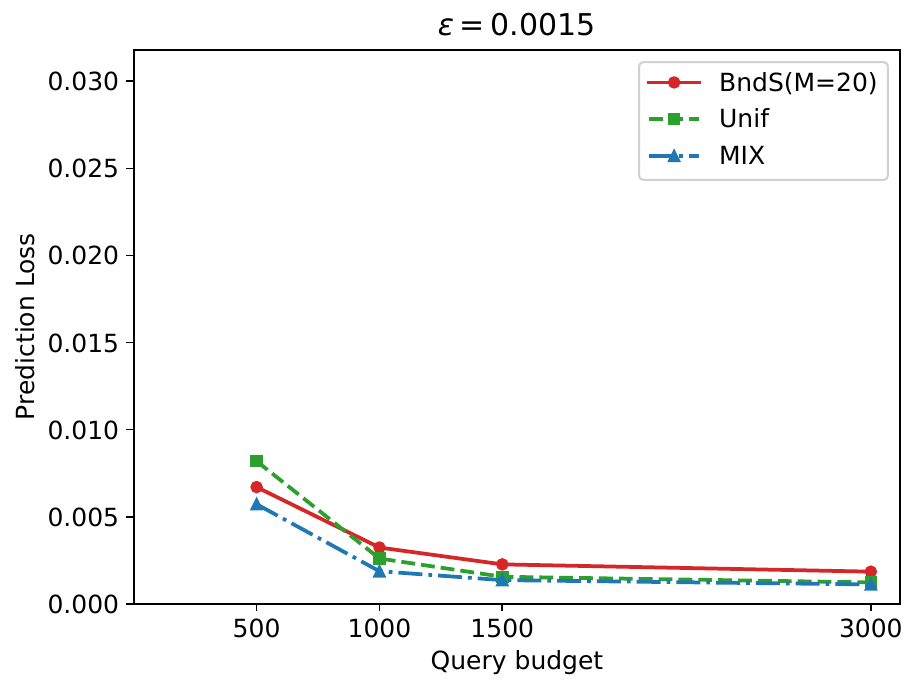}
\end{subfigure}
\caption{Prediction loss under increasing query budgets for BndS ($M=20$), Unif, and MIX, with $\epsilon=0$ (left) and $\epsilon=0.0015$ (right). Curves show medians over five repetitions with $d=10$ and $n=5$.}
\label{fig:budget}
\end{figure*}

Figure~\ref{fig:budget} compares prediction loss under query budgets up to $3000$ for BndS, Unif, and a mixture of the two sampling methods (MIX). MIX uses equal numbers of boundary and uniform samples. For $\epsilon=0$, we use the interface without any perturbation. All three strategies attain near-zero prediction loss at the largest evaluated budget. For $\epsilon=0.0015$, prediction loss decreases as the budget grows, but at $3000$ queries all three strategies have higher median loss than in the undefended setting. MIX has the lowest or near-lowest median loss at the displayed budgets. These results demonstrate finite-budget mitigation for the evaluated attacks, and our defense effectively reduces model extraction accuracy even under large query budget.

\begin{figure*}[!t]
\centering
\begin{subfigure}{0.32\textwidth}
\centering
\includegraphics[width=\textwidth]{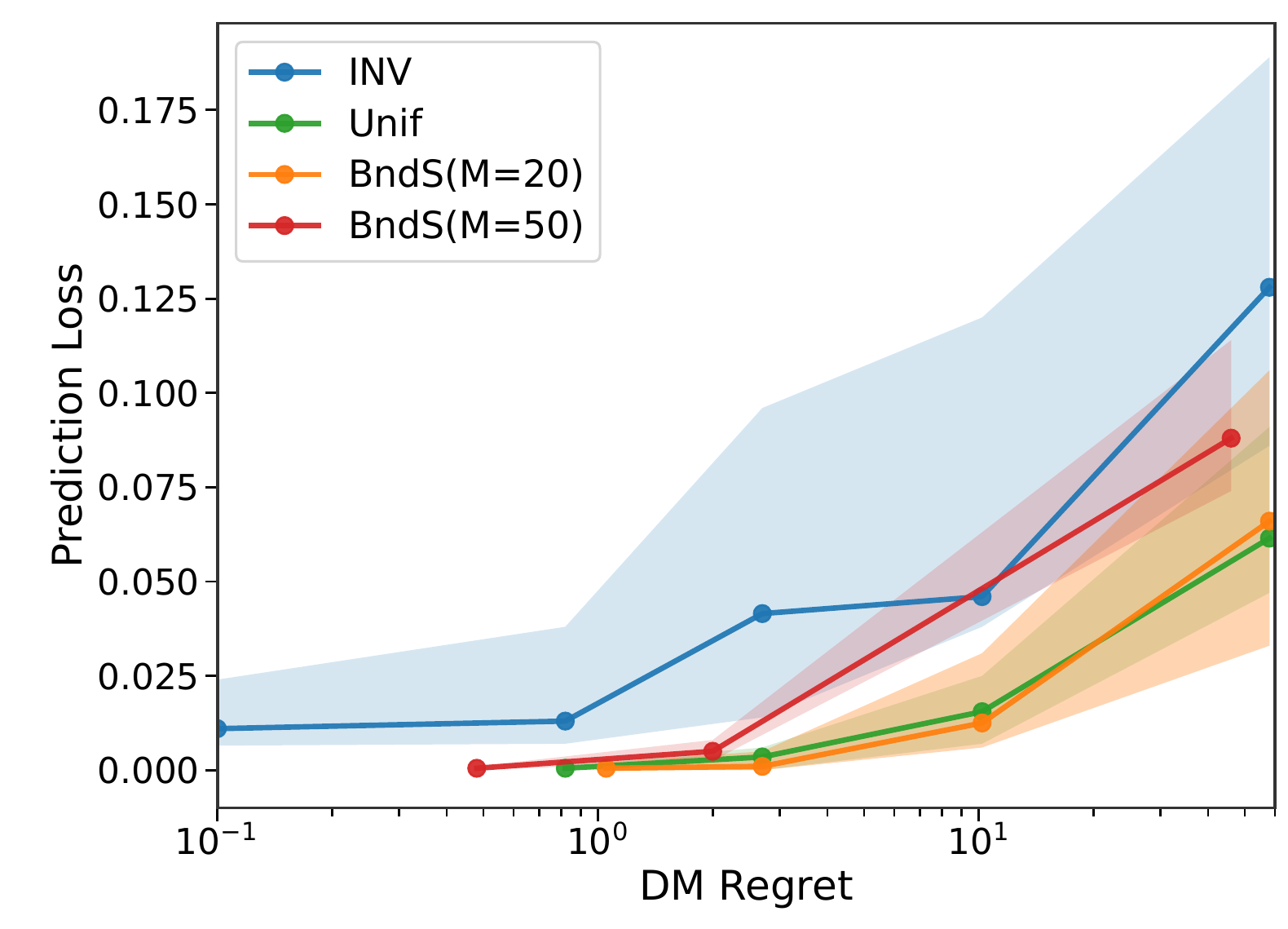}
\caption{}
\label{fig:2a}
\end{subfigure}
\begin{subfigure}{0.32\textwidth}
\centering 
\includegraphics[width=\textwidth]{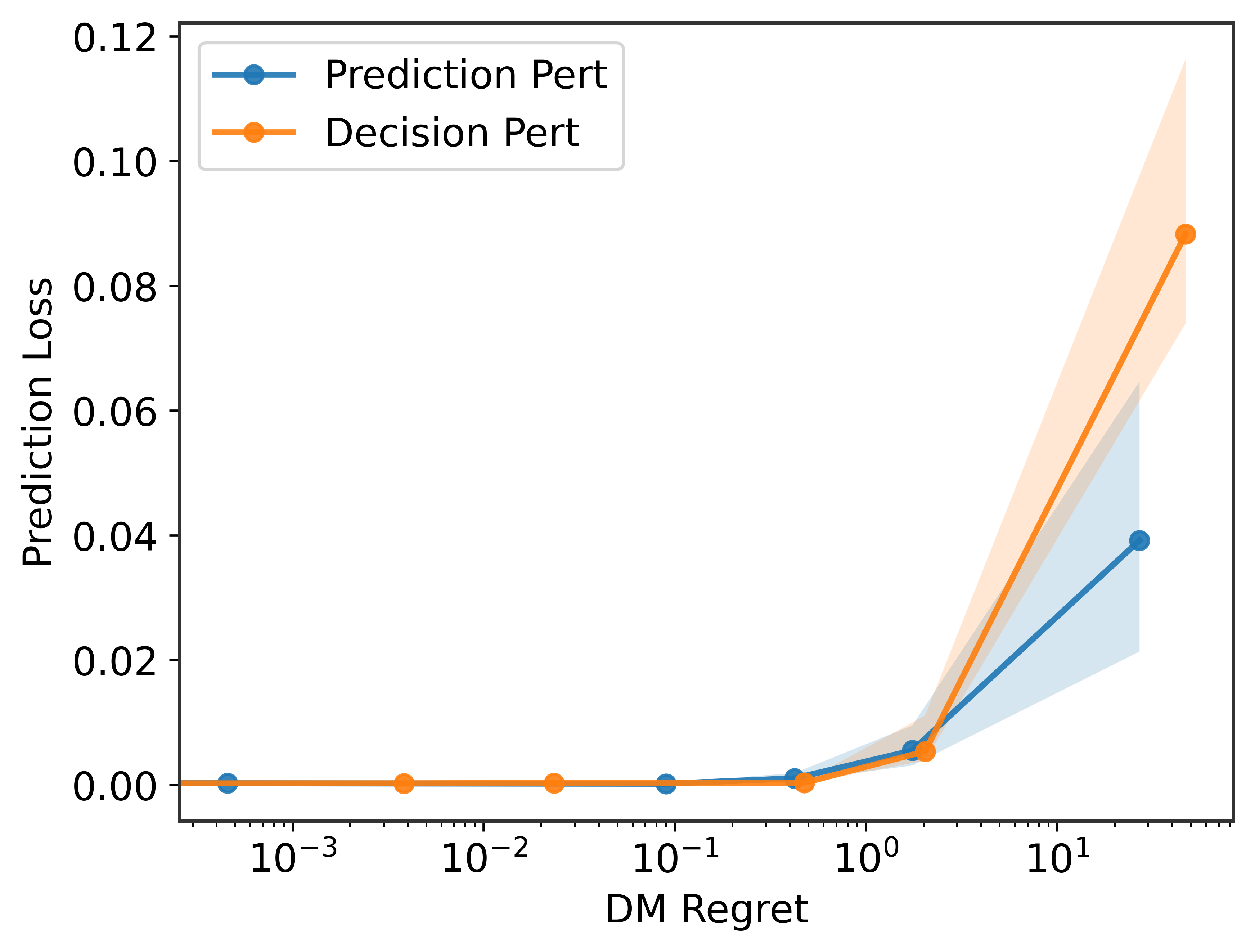}
\caption{}
\label{fig:2b}
\end{subfigure}
\begin{subfigure}{0.32\textwidth}
\centering 
\includegraphics[width=\textwidth]{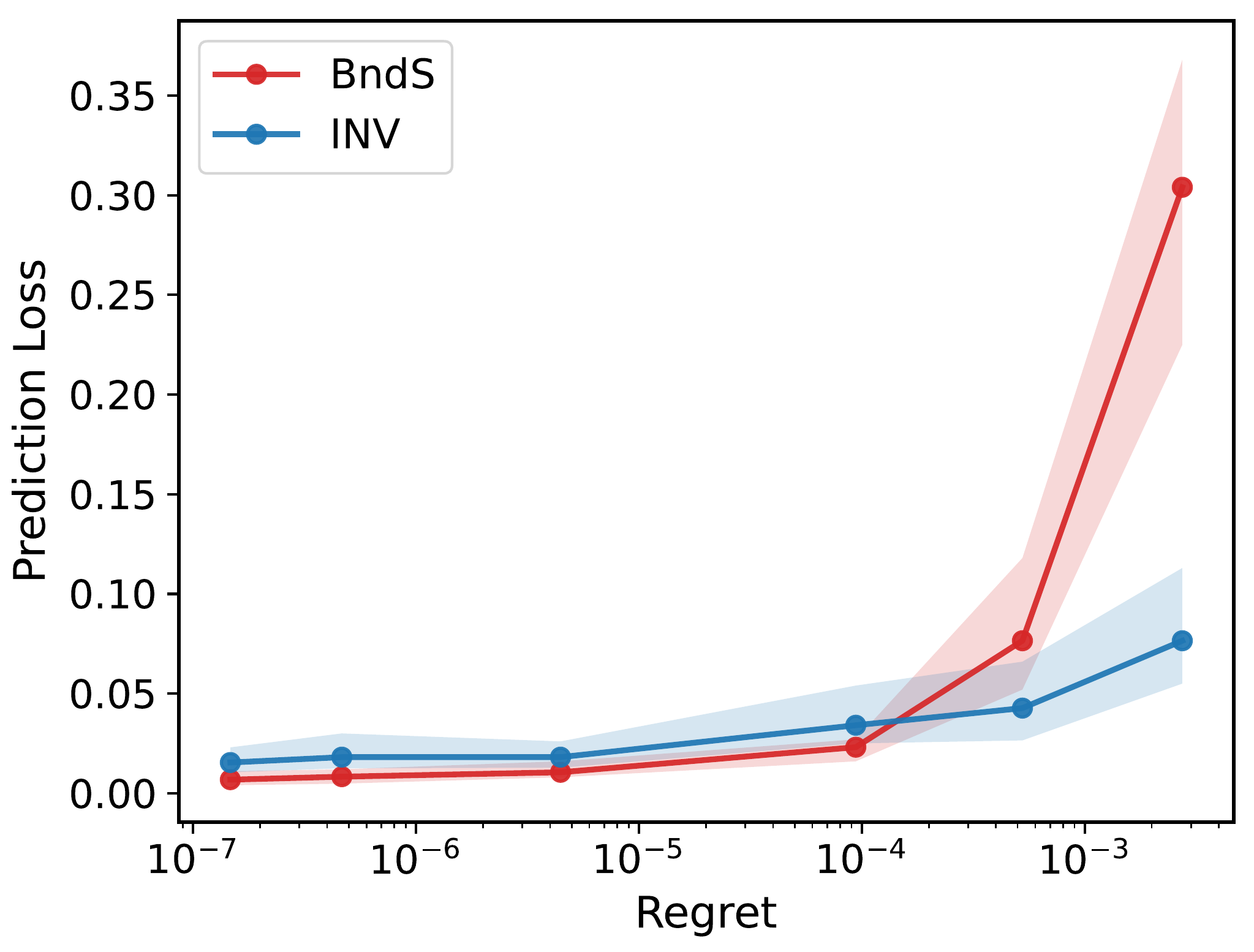} 
\caption{} 
\label{fig:2c}
\end{subfigure}
\caption{Empirical prediction loss-regret trade-offs: (a) decision perturbation against different attacks on synthetic instances; (b) prediction versus decision perturbation on synthetic instances; and (c) decision perturbation on the short-video recommendation task. Points show median regret and median prediction loss, and shaded bands show the $25^\textnormal{th}$-$75^\textnormal{th}$ percentiles of prediction loss over 100 repetitions per parameter setting.}
\label{fig:2}
\end{figure*}

Figure~\ref{fig:2} reports attack-specific prediction loss-regret trade-offs. In particular, Figure~\ref{fig:2a} compares attacks under decision perturbation: the attacker who aims to get more boundary information (BndS(M=50)) suffers from larger prediction under defense. Figure~\ref{fig:2b} shows decision perturbation increases the attacker’s prediction loss at modest regret and is more effective than prediction perturbation. These comparisons concern the reported attacks and parameter settings. 

\textbf{Short video recommendation.} We examine the performance of our framework on a real dataset. We consider a short video recommendation problem constructed from the public dataset KuaiRand-1K \citep{Gao-2022-Kuairand}. We use 800 users for training and 200 for testing, with $d=18$ contextual features and $n=5$ video types. Recommendations satisfy $w \in [0,1]^5$ and $\sum_{i=1}^5w_i \leq 3$. In this experiment, the user contextual information is used to fit a linear predictor of click-through rates, and recommendations are selected by solving a resource-allocation problem using these predictions (see Appendix~\ref{sec:app:exp-setting}). The cost vector in Eq.~\eqref{eq:contextOpt} is the negative of the predicted click-through-rate vector in this example. 

Figure~\ref{fig:2c} shows that BndS has lower prediction loss than INV at the lowest plotted regret levels. At larger regret levels, both attacks incur greater prediction loss, with a more pronounced increase for BndS. This is compatible with the mechanism in Theorem~\ref{thm:DM-def-BS}: making boundary query causes perturbed responses, and fitting perturbed responses as optimal decisions can cause higher prediction loss.

\section{Conclusion}
This paper studies model extraction and defense in contextual linear optimization under the PTO paradigm. We characterize vulnerability through loss calibration, develop an adaptive inverse optimization attack with boundary sampling, and propose a near-boundary decision-perturbation defense. Our theory bounds the regret of the defense and provides a lower bound on expected prediction error for attackers that fit the defended responses as optimal decisions under the evaluation distribution. This lower bound is nondecreasing in the perturbation degree. Experiments show that the defense can increase prediction error
for the evaluated attacks within the tested query budgets. Future work includes analyzing attackers that explicitly model the defense and extending the analysis to nonlinear contextual optimization.

% \subsubsection*{Acknowledgements}

\clearpage
\bibliographystyle{plainnat}
\bibliography{ref}

\clearpage
\appendix
\thispagestyle{empty}

% Supplementary material: To improve readability, you must use a single-column format for the supplementary material.
\onecolumn
\aistatstitle{Supplementary Material: \\ Preventing Model Extraction in Contextual Linear Optimization}

\section{Additional Technical Material}
\subsection{Condition of loss calibration} \label{sec:app:loss-cali}
We present the condition of loss calibration and provide additional insights into how the decision-maker can assess vulnerability. For a full-rank attacker matrix $P^a$, Proposition~\ref{prop:edge-sample} implies that $w^\star(P^{DM}x)\in W^\star(P^ax)$ for every $x \notin \ker(P^{DM})$ ensures positive alignment of $P^a\bar{x}$ and
$P^{DM}\bar{x}$ at every boundary context $\bar{x}$ of $P^a$. Recall that $r_i=-\frac{a_i}{\|a_i\|}$. For a
full-rank attacker matrix $P^a$, we define the candidate space
\begin{equation*}
\mathcal{P}_c(P^a) = \{P \in \R^{n \times d}: P\bar{x} \in \operatorname{span}(r_i) \textnormal{ for every } i=1,2, \ldots, m \textnormal{ and every } \bar{x} \in \R^d \textnormal{ satisfying } P^a\bar{x} = r_i\}.
\end{equation*}
The candidate space is a linear subspace that contains $P^a$. Its definition requires only that the images of the boundary contexts lie on the corresponding lines. In particular, this space includes the zero matrix, although the zero matrix is excluded from the attacker class. 

If $P^a$ has full row rank and satisfies global inverse feasibility, Proposition~\ref{prop:edge-sample} gives
$P^{DM} \in \mathcal{P}_c(P^a)$. If this space is $1$-dimensional, $P^{DM}=\gamma P^a$ for some scalar
$\gamma$. Since $P^{DM} \neq 0$, this scalar is nonzero and the positive boundary alignment in Proposition~\ref{prop:edge-sample} implies $\gamma>0$. Thus the two matrices have zero prediction loss at every context. Then, we connect Assumption~\ref{assumption:main} to the dimension of $\mathcal{P}_c(P^a)$ by the following result:
\begin{proposition} \label{prop:1-dim-suff-condition}
Suppose that Assumption~\ref{assumption:main} holds. Then, for every full-rank matrix $P^a \in \R^{n \times d}$, we have $\mathcal{P}_c(P^a)=\operatorname{span}\{P^a\}$. In particular, the dimension of $\mathcal{P}_c(P^a)$ is $1$.
\end{proposition}
Proposition~\ref{prop:1-dim-suff-condition} ensures that $\mathcal{P}_c(P^{DM})$ is $1$-dimensional under Assumption~\ref{assumption:main}. The following corollary shows that this property is sufficient for
loss calibration. 
\begin{corollary} \label{coro:calibrate-condition-DM}
If $\mathcal{P}_c(P^{DM})$ is $1$-dimensional, every inverse feasibility loss calibrates the prediction loss.
\end{corollary}
Corollary~\ref{coro:calibrate-condition-DM} provides a diagnostic for the decision-maker. For each facet $i$, we choose $d-n+1$ linearly independent contexts $\{\bar{x}_{i, j}\}_{j=1}^{d-n+1}$ satisfying $P^{DM}\bar{x}_{i, j}=r_i$ for all $j = 1, 2, \ldots, d-n+1$ and set $X_i=[\bar{x}_{i, 1},\ldots,\bar{x}_{i, d-n+1}]$.
Since $P^{DM}$ has full row rank, the linear space $H_i=\{x \in \R^d: P^{DM}x \in \operatorname{span}(r_i)\}$ has dimension $d-n+1$, and the columns of $X_i$ form a basis of $H_i$. Such representatives exist: if $y_i$ satisfies $P^{DM}y_i=r_i$ and $z_1,\ldots,z_{d-n}$ form a basis of $\ker(P^{DM})$, one may take $y_i,y_i+z_1,\ldots,y_i+z_{d-n}$.

Since $r_i$ is a unit vector, the condition $P\bar{x}_{i, j} \in \operatorname{span}(r_i)$ is equivalent
to $(I_n-r_ir_i^\top)P\bar{x}_{i, j}=0$. This implies 
\begin{equation*}
\mathcal{P}_c(P^{DM}) = \{P \in \R^{n \times d}: (I_n-r_ir_i^\top)PX_i=0 \textnormal{ for every } i = 1, 2, \ldots, m\}.
\end{equation*}
Thus, the decision-maker can determine the dimension of the candidate space by solving a linear system in the entries of $P$. If its solution space is $1$-dimensional, Corollary~\ref{coro:calibrate-condition-DM} implies loss calibration. This diagnostic uses $P^{DM}$ and is available to the decision-maker.

\subsection{Inverse optimization methods} \label{sec:app-io-loss-opt}
We present other loss functions motivated by inverse optimization and the approaches to solve them. Any inverse feasibility loss function can be incorporated into our framework, although the resulting empirical optimization problems are different and may require different solution methods.

\paragraph{Scale normalization of the suboptimality loss function.} The suboptimality-loss formulation in Eq.~\eqref{eq:subopt-reform} is homogeneous and admits $P=0$. In the experiments, we solve two convex programs with the respective constraints
\begin{equation*}
\sum_{i=1}^n \sum_{j=1}^d P_{ij} \geq 1 \textnormal{ or } \sum_{i=1}^n \sum_{j=1}^d P_{ij} \leq -1.
\end{equation*}
Each constraint excludes the zero matrix. Every matrix with a nonzero entry sum has a positive rescaling satisfying one of these constraints, so zero-loss feasibility is preserved for such matrices. Matrices with zero entry sum are excluded. Since positive rescaling changes the value of the suboptimality loss, the fitted matrix is normalized in Frobenius norm before applying the stopping test in Algorithm~\ref{algo:attack}. 

\paragraph{Distance loss function.} The distance loss measures the distance from an observed decision
to the optimal solution set under $P$: 
\begin{equation}\label{eq:dist-loss}
L_d(P, P^{DM}; x_i) = \min_{w \in W^\star(Px_i)} \|w - w_i\|.
\end{equation}
We define the empirical distance loss by
\begin{equation*}
\hat{L}_d(P,P^{DM};D) = \sum_{i=1}^N L_d(P,P^{DM};x_i).
\end{equation*}
Recall that the set $W^\star(Px_i)$ is nonempty and compact. Thus, the minimum in Eq.~\eqref{eq:dist-loss} is attained and we have $L_d(P,P^{DM};x_i)=0$ if and only if $w_i \in W^\star(Px_i)$. By nonnegativity, the empirical loss vanishes exactly on $\mathcal{P}_\textnormal{inv}(D)$ when restricted to nonzero matrices. Since $W^\star(0)=S$, this also holds when $Px_i=0$. 

The distance loss is not continuous in general. Following the duality-based approach of~\citet{Aswani-2018-Inverse}, we consider an optimality-gap relaxation for the fixed feasible set $S$.
For $\epsilon \geq 0$, we define
\begin{equation*}
\begin{array}{rcl}
\hat{L}_d(P,P^{DM};D,\epsilon) = & \min_{\{\tilde{w}_i,\lambda_i\}_{i=1}^N} & \sum_{i=1}^N \|w_i-\tilde{w}_i\|,\\
& \textnormal{s.t.} & A^\top\lambda_i=Px_i, \ i=1,\ldots,N,\\
& & (Px_i)^\top\tilde{w}_i-b^\top\lambda_i \leq \epsilon, \ i=1,\ldots,N,\\
& & A\tilde{w}_i \leq b, \ i=1,\ldots,N,\\
& & \lambda_i \leq 0, \ i=1,\ldots,N.
\end{array}
\end{equation*}
Strong linear programming duality makes these constraints equivalent to $(Px_i)^\top\tilde{w}_i-z^\star(Px_i) \leq \epsilon$ and $\tilde{w}_i \in S$ for every $i=1, 2, \ldots, N$. This implies $\hat{L}_d(P,P^{DM};D,0)=\hat{L}_d(P,P^{DM};D)$. For $\epsilon>0$, the relaxed loss is continuous in $P$ but need not be an inverse feasibility loss. For every fixed $P$, we have 
\begin{equation*}
\hat{L}_d(P,P^{DM};D,\epsilon) \uparrow \hat{L}_d(P,P^{DM};D) \textnormal{ as } \epsilon \downarrow 0.
\end{equation*}
When minimization is over a fixed nonempty compact set of nonzero matrices, every accumulation point of global minimizers of the relaxed loss as $\epsilon \downarrow 0$ minimizes the exact empirical distance loss over the same set.

\paragraph{CaVE loss function.} Following the approach of~\citet{Tang-2024-Cave}, we define the cone-aligned vector estimation (CaVE) loss function as follows, 
\begin{equation}\label{eq:loss}
L_C(P^a,P^{DM};x_i)=
\begin{cases}
1-\cos(\angle(P^ax_i,\cone(w_i))), & \textnormal{if } P^ax_i \neq 0,\\
0,& \textnormal{otherwise}.
\end{cases}
\end{equation}
Its empirical value is
\begin{equation*}
L_C(P^a,P^{DM};D) = \tfrac{1}{N} \sum_{i=1}^N L_C(P^a,P^{DM};x_i).
\end{equation*}
For nonzero costs, zero angular distance to the closed normal cone is equivalent to being in that cone. At zero cost, every feasible decision is optimal. For every nonzero matrix $P^a$, we have
\begin{equation*}
L_C(P^a,P^{DM};D)=0 \Longleftrightarrow w_i \in W^\star(P^ax_i) \textnormal{ for all } i=1,\ldots,N.
\end{equation*}
Thus, $L_C$ is an inverse feasibility loss. The loss can be discontinuous at a zero predicted cost.

For each observed vertex $w_i$, we let $I_i=\{j:a_j^\top w_i=b_j\}$ and let $A_{I_i}$ contain the corresponding rows of $A$. Then, we have $\cone(w_i)=\{-A_{I_i}^\top\lambda_i:\lambda_i \geq 0\}$. Here, $\lambda_i \geq 0$ contains the coefficients of the inward active normals $-a_j$ for $j \in I_i$. For $P^ax_i \neq 0$, its angular loss can therefore be written as
\begin{equation*}
L_C(P^a,P^{DM};x_i) = \min_{\substack{\lambda_i \geq 0 \\ \|A_{I_i}^\top\lambda_i\|=1}}\left(1+\tfrac{(A_{I_i}^\top\lambda_i)^\top P^ax_i}{\|P^ax_i\|}\right).
\end{equation*}
With a numerical bound $C \geq \frac{1}{nd}$, we minimize the empirical loss subject to an entry-sum normalization and bounds on the matrix entries:
\begin{equation}\label{eq:loss-reformulation1}
\begin{array}{rcl}
& \min_{P^a,\lambda_1,\ldots,\lambda_N} & \tfrac{1}{N} \sum_{i=1}^N\left(1+\tfrac{(A_{I_i}^\top\lambda_i)^\top P^ax_i}{\|P^ax_i\|}\right) \\
& \textnormal{s.t.} & \lambda_i \geq 0,\ \|A_{I_i}^\top\lambda_i\|=1, \ i=1,\ldots,N,\\
& & \sum_{i=1}^n \sum_{j=1}^d P^a_{ij}=1,\\
& & |P^a_{ij}| \leq C, \ i=1,\ldots,n, \ j=1,\ldots,d.
\end{array}
\end{equation}
In this formulation and the expressions below, the entire angular summand is defined to be zero when its predicted cost is zero, as in Eq.~\eqref{eq:loss}. The matrix constraints restrict the admissible predictors. 

Introducing a local matrix $P_i$ for each observation gives the consensus formulation as follows, 
\begin{equation} \label{eq:loss-reformulation2}
\begin{array}{rcl}
& \min_{P^a,\{P_i,\lambda_i\}_{i=1}^N} & \tfrac{1}{N} \sum_{i=1}^N\left(1+\tfrac{(A_{I_i}^\top \lambda_i)^\top P_ix_i}{\|P_ix_i\|}\right) \\
& \textnormal{s.t.} & \lambda_i \geq 0,\ \|A_{I_i}^\top\lambda_i\|=1, \ i=1,\ldots,N,\\
& & P_i=P^a, \ i=1,\ldots,N,\\
& & \sum_{i=1}^n \sum_{j=1}^d P^a_{ij}=1,\\
& & |P^a_{ij}| \leq C, \ i=1,\ldots,n,\ j=1,\ldots,d.
\end{array}
\end{equation}
For $\rho>0$, the augmented Lagrangian function is defined as follows, 
\begin{equation*}
L_\rho(\{P_i\},\{\lambda_i\},P^a,\{\mu_i\}) = \sum_{i=1}^N\left[\tfrac{1}{N}\left(1+\tfrac{(A_{I_i}^\top \lambda_i)^\top P_ix_i}{\|P_ix_i\|}\right)+\langle\mu_i,P_i-P^a\rangle_F+\tfrac{\rho}{2}\|P_i-P^a\|_F^2
\right],
\end{equation*}
where $\mu_i \in \R^{n \times d}$ is the multiplier for $P_i=P^a$. The ADMM method in Algorithm~\ref{algo:ADMM} alternates between updates of $P^a$, $\lambda_i$, and $P_i$, followed by the multiplier
updates. In the local updates of Algorithm~\ref{algo:ADMM}, $L_\rho(P_i,\lambda_i,P^a,\mu_i)$ denotes the $i^\textnormal{th}$ summand of the augmented Lagrangian function. We also write $\Lambda_i=\{\lambda_i \in \R^{|I_i|}: \lambda_i \geq 0, \|A_{I_i}^\top\lambda_i\|=1\}$. 
\begin{algorithm}[!t] 
\caption{Learning the prediction matrix via ADMM} \label{algo:ADMM}
\begin{algorithmic}[1]
\STATE \textbf{Input:} $P^{a,0}$ satisfying $\sum_{i=1}^n \sum_{j=1}^d P^{a,0}_{ij}=1$ and $|P^{a,0}_{ij}|\leq C$ for all $i,j$, $\rho>0$, and $T\geq 1$.
\STATE \textbf{Initialize:} $t=0$ and $P_i^0=P^{a,0}$, $\mu_i^0=0$, and $\lambda_i^0=\frac{\mathbf{1}}{\|A_{I_i}^\top\mathbf{1}\|}$ for $i=1,\ldots,N$. 
\WHILE{$t < T$} 
\STATE $P^{a,t+1} \in \argmin_{\sum_{i=1}^n \sum_{j=1}^d P_{ij}=1, |P_{ij}| \leq C \textnormal{ for all }i, j}
L_\rho(\{P_i^t\},\{\lambda_i^t\},P,\{\mu_i^t\})$. 
\FOR{$i \in \{1, \ldots, N\}$}
\STATE $\lambda_i^{t+1} \in \argmin_{\lambda_i \in \Lambda_i} L_\rho(P_i^t, \lambda_i, P^{a,t+1}, \mu_i^t)$.
\STATE $P_i^{t+1} \in \argmin_{P_i} L_\rho(P_i, \lambda_i^{t+1}, P^{a,t+1}, \mu_i^t)$.  
\STATE $\mu_i^{t+1} = \mu_i^t + \rho(P_i^{t+1} - P^{a,t+1})$.  
\ENDFOR 
\STATE $t \gets t + 1$.  
\ENDWHILE 
\STATE \textbf{Return:} $P^{a,t}$.
\end{algorithmic}
\end{algorithm}

The local minimizations can be approximated by first-order updates. For $P_i^tx_i\ne0$, a projected gradient step for $\lambda_i$ and a gradient step for $P_i$ are
\begin{equation}
\begin{array}{rcl}
\lambda_i^{t+1} & \in & \Pi_{\Lambda_i}\left(\lambda_i^t-\tfrac{\alpha}{N}\tfrac{A_{I_i}P_i^tx_i}{\|P_i^tx_i\|}
\right),\\
P_i^{t+1} & = & P_i^t-\beta\left(\tfrac{1}{N}\left(\tfrac{A_{I_i}^\top\lambda_i^{t+1}}{\|P_i^tx_i\|}-\tfrac{(A_{I_i}^\top\lambda_i^{t+1})^\top P_i^tx_i}{\|P_i^tx_i\|^3}P_i^tx_i\right)x_i^\top+\mu_i^t+\rho(P_i^t-P^{a,t+1})\right),
\end{array}
\end{equation}
where $\alpha,\beta>0$ are step sizes and $\Pi_{\Lambda_i}$ denotes Euclidean projection onto the feasible set $\Lambda_i$. The projection may have multiple solutions, in which case any minimizer can be selected. These updates require nonzero local predicted costs and ADMM is used as a numerical method without a global convergence guarantee.

\begin{figure}[!t]
\centering
\begin{subfigure}[t]{0.48\textwidth}
\centering 
\includegraphics[width=\textwidth]{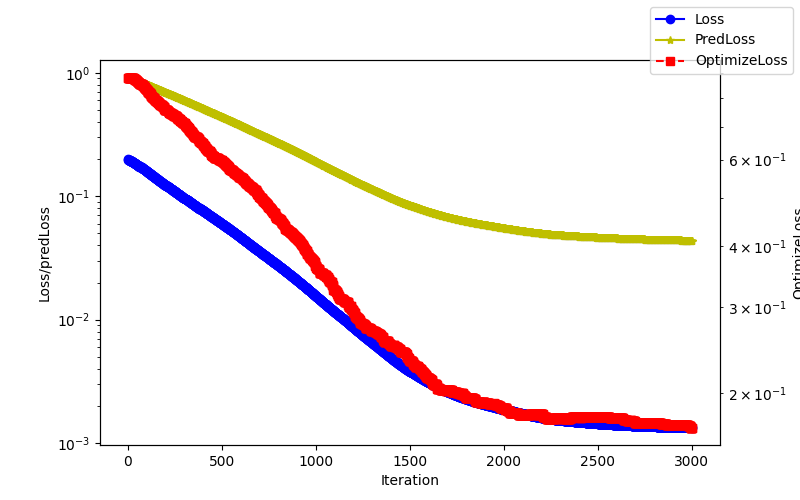} \caption{ADMM} 
\label{subfig:convergence-admm}
\end{subfigure}
\begin{subfigure}[t]{0.48\textwidth}
\centering
\includegraphics[width=\textwidth]{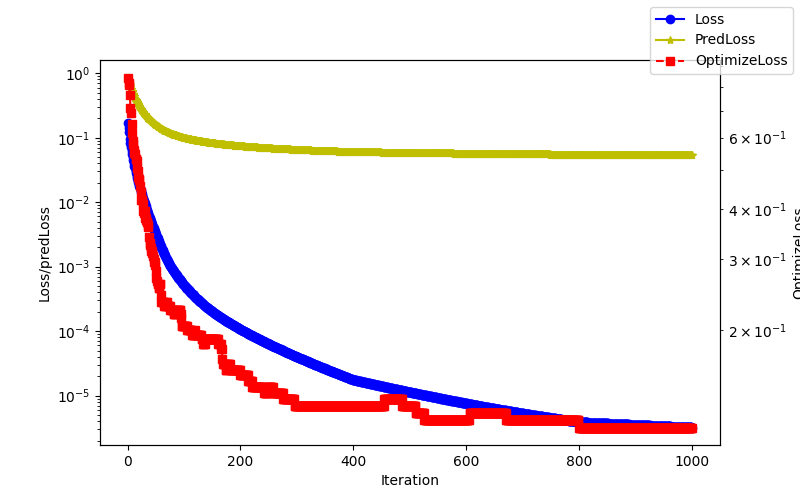}
\caption{Tang and Khalil's method}
\label{subfig:convergence-heur}
\end{subfigure}
\caption{Convergence of the ADMM algorithm and Tang and Khalil's method. The red line represents the optimization loss, defined as the proportion of incorrect decisions $w^\star(P^a x_i) \neq w^\star(P^{DM}x_i)$ over the dataset.}
\label{fig:convergence}
\end{figure}

The experiment shows that our solution approach can achieve near-zero optimality gap. Apart from ADMM, we tried the heuristic algorithm~\citep{Tang-2024-Cave} and the global solver in scipy package to solve the optimization problem. Figure \ref{fig:convergence} shows the convergence of ADMM and Tang and Khalil's method. We observe that both methods achieve near zero CaVE loss. Table \ref{tab:algorithm-comparison} compares all methods and shows that the global solver attains the smallest reported CaVE loss. Furthermore, if we encounter some context such that $\|P^{a}x_i\| = 0$, then we discard this context in practice.
\begin{table}[!t]
\centering
\caption{Comparison of three approaches for minimizing the CaVE loss.} \label{tab:algorithm-comparison}
\begin{tabular}{lccccc} \toprule
\textbf{Algorithm} & \textbf{CaVE loss} & \textbf{predLoss} & \textbf{optLoss} & \\ \midrule
ADMM & 0.0007018 & 0.0356 & 0.1450 \\
Tang and Khalil's method & 0.0000032 & 0.0539 & 0.1150 \\
Global & 0.0000030 & 0.0378 & 0.0650 \\ \bottomrule
\end{tabular}
\end{table}

\subsection{Sampling algorithms}\label{sec:app:sampling}
We specify the sampling step of Algorithm~\ref{algo:attack}. Indeed, for a full-rank matrix $P^a$, the uniform sampling method in Algorithm~\ref{algo:unif-sampling} draws a direction uniformly from the unit sphere in $\R^n$ and chooses a context whose predicted cost has that direction. Since the context is chosen using $P^a$, the sampling rule is adaptive, unlike a nonadaptive batch of contexts used by the inverse optimization benchmark. Uniform sampling is simple, and its sample size controls the number of observations added at each iteration. It does not target the decision boundaries of the downstream problem. To exploit this geometry, we use the boundary sampling method in Algorithm~\ref{algo:edge-sampling}.

\begin{figure}[!ht]
\centering
\resizebox{0.7\textwidth}{!}{%
\begin{circuitikz}
\tikzstyle{every node}=[font=\normalsize]
\draw [short] (12.5,18.75) -- (15,18.75);
\draw [short] (15,18.75) -- (16.25,16.25);
\draw [->, >=Stealth] (15,18.75) -- (15,21.75);
\draw [->, >=Stealth] (15,18.75) -- (17.5,20);
\draw [short] (11.25,16.25) -- (12.5,18.75);
\draw [short] (11.25,16.25) -- (16.25,16.25);
\node [font=\normalsize] at (15,22) {$a_i$};
\draw [->, >=Stealth] (12.5,18.75) -- (12.5,21.75);
\node [font=\normalsize] at (12.5,22) {$a_i$};
\draw [->, >=Stealth] (12.5,18.75) -- (10,20);
\node [font=\normalsize] at (16.25,20.5) {$C_k, k\in \mathcal{C}_i$};
\node [font=\normalsize] at (11.5,20.5) {$C_j, j\in\mathcal{C}_i$};
\draw [short] (3.75,18.75) -- (6.25,18.75);
\draw [short] (6.25,18.75) -- (7.5,16.25);
\draw [->, >=Stealth] (6.25,18.75) -- (6.25,21.75);
\draw [->, >=Stealth] (6.25,18.75) -- (8.75,20);
\draw [short] (2.5,16.25) -- (3.75,18.75);
\draw [short] (2.5,16.25) -- (7.5,16.25);
\node [font=\normalsize] at (6,18.5) {$w_k$};
\node [font=\normalsize] at (7,20.25) {$C_k$};
\node [font=\normalsize] at (6.25,22) {$a_i,i\in I_k$};
\node [font=\normalsize] at (9,20.25) {$a_j,j\in I_k$};
\node [font=\normalsize] at (12.75,18.5) {$w_j$};
\node [font=\normalsize] at (14.75,18.5) {$w_k$};
\node [font=\normalsize] at (5,15.75) {(a)};
\node [font=\normalsize] at (13.75,15.75) {(b)};
\end{circuitikz}
}%

\caption{(a) the tight constraints (also the boundaries of normal cone) incident to the a normal cone $C_k$. (b) the normal cones (also the vertex) incident to the constraint $a_i$.}
\label{fig:poly}
\end{figure}
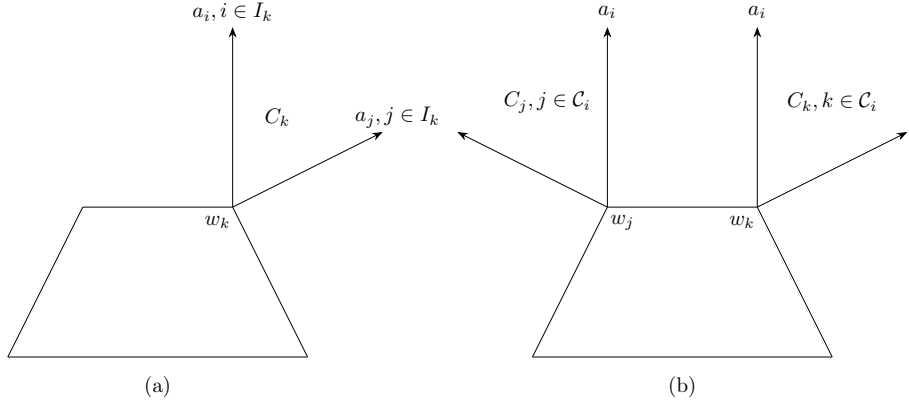

We enumerate the vertices as $V = \{w_1,\ldots,w_{|V|}\}$ and write $C_k=\cone(w_k)$. For each vertex $w_k \in V$, we let $I_k=\{i \in \{1,\ldots,m\}: a_i^\top w_k=b_i\}$. Then, for each facet $i$, we define the set of incident vertex indices by $\mathcal{C}_i=\{k \in \{1,\ldots,|V|\}:i\in I_k\}$ and set $K=\max_{1 \leq k \leq |V|} |I_k|$. Using the inward directions $r_i=-\frac{a_i}{\|a_i\|}$, we have
\begin{equation*}
C_k=\operatorname{polycone}\{r_i:i\in I_k\}, \quad \bigcap_{k \in \mathcal{C}_i} C_k = \{\alpha r_i: \alpha \geq 0\}.
\end{equation*}
Figure~\ref{fig:poly} illustrates the facet-vertex incidences. For each facet $i$, Algorithm~6 chooses $d-n+1$ linearly independent boundary contexts satisfying $P^a\bar{x}_{i0j}=r_i$ for all $j=1,\ldots,d-n+1$
and sets $X_i=[\bar{x}_{i01},\ldots,\bar{x}_{i0,d-n+1}]$. These columns span the $(d-n+1)$-dimensional linear space $\{x \in \R^d: P^ax \in \operatorname{span}(r_i)\}$. For $M \geq 1$, each $k \in \mathcal{C}_i$, and each $j=1,\ldots,d-n+1$, the associated near-boundary context is
\begin{equation*}
\tilde{x}_{ikj} = \tfrac{M\bar{x}_{i0j}+\sum_{\ell \in I_k \setminus \{i\}} \bar{x}_{\ell0j}}{\|Mr_i+\sum_{\ell \in  I_k \setminus \{i\}} r_\ell\|}.
\end{equation*}
The vector in the denominator is a positive combination of all active inward normals at $w_k$, so it belongs to $\operatorname{int}(C_k)$ and is nonzero. Thus, we have
\begin{equation*}
P^a\tilde{x}_{ikj} = \tfrac{Mr_i+\sum_{\ell \in I_k \setminus \{i\}} r_\ell}{\|Mr_i+\sum_{\ell \in I_k \setminus \{i\}} r_\ell\|} \in \operatorname{int}(C_k).
\end{equation*}
Thus, $w_k$ is the unique optimizer under the attacker cost at $\tilde{x}_{ikj}$, and $P^a\tilde{x}_{ikj} \to r_i$ as $M \to +\infty$. The algorithm queries all boundary contexts and all of these near-boundary contexts.

In what follows, we describe the stopping condition mentioned in Theorem~\ref{thm:sample-global-error-informal}. Indeed, we define 
\begin{equation*}
\eta_S = \min_{1 \leq k \leq |V|} \min_{v\in V\setminus\{w_k\}} \sum_{\ell \in I_k} r_\ell^\top(v-w_k).
\end{equation*}
Clearly, $\eta_S \geq 0$. For $v \neq w_k$, at least one is strictly positive, because the active facet normals at $w_k$ span $\R^n$. Thus, $\eta_S>0$. For $M \geq 1$, $k \in \mathcal{C}_i$, and $v \neq w_k$, we have
\begin{equation*}
(P^a\tilde{x}_{ikj})^\top(v-w_k) = \tfrac{(M-1)r_i^\top(v-w_k)+\sum_{\ell \in I_k} r_\ell^\top(v-w_k)}
{\|Mr_i+\sum_{\ell \in I_k \setminus \{i\}} r_\ell\|} \geq \tfrac{\eta_S}{M+K-1}.
\end{equation*}
We impose the sufficient stopping condition as follows, 
\begin{equation}\label{eq:stopping}
\Delta(M+K-1)<\eta_S.
\end{equation}
If a near-boundary query returns any vertex other than $w_k$, its suboptimality loss under $P^a$ then exceeds $\Delta$, so Algorithm~\ref{algo:attack} cannot terminate at that sampling call.

\begin{theorem} \label{thm:sample-global-error} 
Suppose Assumption~\ref{assumption:main} holds. For each $\Delta>0$ in a family tending to zero, run Algorithm~\ref{algo:attack} with a full-rank initialization and Algorithm~\ref{algo:edge-sampling} with $M = M(\Delta) \geq 1$. Assume that every inverse optimization update returns a full-rank matrix with zero inverse feasibility loss on the accumulated dataset. Suppose there exist constants $R>0$ and $\kappa>0$, independent of the iteration and of $\Delta$, such that every queried context satisfies $\|x\| \leq R$ and $\sigma_{\min}(X_i) \geq \kappa$ for all $i=1,\ldots,m$ at every sampling call. Here, $\sigma_{\min}(X_i)$ denotes the smallest of the $d-n+1$ singular values of $X_i$. Assume that $M(\Delta) \to +\infty$ as $\Delta \downarrow 0$ and that Eq.~\eqref{eq:stopping} holds for all sufficiently small $\Delta$. Then, every run terminates after at most $\left(1+\frac{2RL_e}{\Delta}\right)^{nd}$ inverse optimization updates and its normalized output satisfies $P^a_{M(\Delta),\Delta} \rightarrow \frac{P^{DM}}{\|P^{DM}\|_F}$. As a consequence, we have $L_\textnormal{pred}(P^a_{M(\Delta),\Delta},P^{DM},x) \rightarrow 0$ for every $x \notin \ker(P^{DM})$. 
\end{theorem}
The bound on $\|x\|$ yields a positive lower bound on the distance between distinct nonterminal iterates, which ensures finite termination. And the uniform lower bound $\sigma_{\min}(X_i)\ge\kappa>0$ ensures that the columns of $X_i$ remain linearly independent in the limit. A sufficient parameter choice is $M(\Delta) \to \infty$ and $\Delta M(\Delta) \to 0$ as $\Delta \downarrow 0$. 
Indeed, these limits imply $\Delta(M(\Delta)+K-1) \to 0$, so Eq.~\eqref{eq:stopping} holds eventually. For example, $M(\Delta)=\max\{1,\Delta^{-1/2}\}$ satisfies these conditions. Theorem~\ref{thm:sample-global-error} concerns this coupled limit. A smaller $\Delta$ increases the stated upper bound on the number of inverse optimization updates.
\begin{algorithm}[!t] \small
\caption{Uniform sampling (Unif)} \label{algo:unif-sampling} 
\begin{algorithmic} 
\STATE \textbf{Input:} A full-rank matrix $P^a$ and a positive integer $N_t$.
\STATE \textbf{Initialize:} $D \leftarrow \varnothing$.
\FOR{$k=1...N_t$} 
\STATE Sample $u \sim \mathcal{N}(0,I_n)$ and set $u \leftarrow \frac{u}{\|u\|}$.
\STATE Choose $x \in \R^d$ satisfying $P^ax=u$.
\STATE Query $x$ and observe $w=w^\star(P^{DM}x)$.
\STATE $D \leftarrow D \cup \{(x,w)\}$.
\ENDFOR 
\STATE \textbf{Return:} $D$. 
\end{algorithmic}
\end{algorithm}
\begin{algorithm}[!t] \small
\caption{Boundary sampling (BndS)} \label{algo:edge-sampling}
\begin{algorithmic}
\STATE \textbf{Input:} A full-rank matrix $P^a$ and a parameter $M \geq 1$.
\STATE \textbf{Initialize:} $D \leftarrow \varnothing$.
\FOR{$i=1, \ldots, m$}
\STATE Choose linearly independent $\bar{x}_{i01},\ldots,\bar{x}_{i0,d-n+1}$ satisfying $P^a\bar{x}_{i0, j}=r_i$ for all $j=1,\ldots,d-n+1$.
\FOR{$j=1, \ldots,d-n+1$}
\STATE Query $\bar x_{i0j}$ and observe $\bar{w}_{i0j}=w^\star(P^{DM}\bar{x}_{i0j})$.
\STATE $D \leftarrow D \cup \{(\bar{x}_{i0j},\bar{w}_{i0j})\}$.
\ENDFOR
\ENDFOR
\FOR{$i=1, \ldots,m$} 
\FOR{$j=1,\ldots,d-n+1$}
\FOR{$k \in \mathcal{C}_i$}
\STATE $\tilde{x}_{ikj} \leftarrow \frac{M\bar{x}_{i0j}+\sum_{\ell \in I_k \setminus \{i\}} \bar{x}_{\ell0j}}{\|Mr_i+\sum_{\ell \in I_k \setminus \{i\}} r_\ell\|}$.
\STATE Query $\tilde{x}_{ikj}$ and observe $\tilde{w}_{ikj}=w^\star(P^{DM}\tilde{x}_{ikj})$. 
\STATE $D \leftarrow D \cup \{(\tilde{x}_{ikj},\tilde{w}_{ikj})\}$. 
\ENDFOR
\ENDFOR
\ENDFOR 
\STATE \textbf{Return:} $D$.  
\end{algorithmic}
\end{algorithm}

\subsection{Restricted context region}
We consider the undefended interface with query contexts restricted to a set $\mathcal{X} \subseteq \R^d$. The following result gives a sufficient condition for identifying the prediction matrix from decisions observed in this region.
\begin{lemma}\label{lem:1-dim-suff-condition-restrict-region}
Let $J \subseteq \{1,\ldots,m\}$ be a set of facet indices such that $\{a_i:i\in J\}$ contains $n+1$ normals satisfying Assumption~\ref{assumption:main}. Suppose that, for each $i \in J$, there exists $\bar{x}_i \in \operatorname{int}(\mathcal{X})$ with $P^{DM}\bar{x}_i=\alpha_i r_i$ for some $\alpha_i>0$. If a nonzero matrix $P^a$ satisfies $w^\star(P^{DM}x) \in W^\star(P^ax)$ for every $x \in \mathcal{X} \setminus \ker(P^{DM})$, $P^a=\gamma P^{DM}$ for some $\gamma>0$.
\end{lemma}
For boundary sampling, fix a set $J \subseteq \{1,\ldots,m\}$ of facet indices. At each sampling call, choose the boundary representatives in Algorithm~\ref{algo:edge-sampling} for all facets, but query only $\bar{x}_{i0j}$ and $\tilde{x}_{ikj}$ for all $i \in J$, $j=1,\ldots,d-n+1$ and $k \in \mathcal{C}_i$. Every query in these selected blocks is retained and must lie in $\mathcal{X}$. Boundary representatives for facets
outside $J$ may still be needed to form the near-boundary mixtures. 

\begin{lemma} \label{lem:restricted-boundary-sampling-convergence}
Suppose $\{a_i:i\in J\}$ contains $n+1$ normals satisfying Assumption~\ref{assumption:main}. For each $\Delta>0$ in a family tending to zero, run Algorithm~\ref{algo:attack} with a full-rank initialization and the restricted sampling rule above, using $M = M(\Delta) \geq 1$. Assume every inverse optimization update returns a full-rank matrix with zero inverse feasibility loss on the accumulated dataset. Suppose there exist constants $R>0$ and $\kappa>0$, independent of the iteration and of $\Delta$, such that every queried context and every boundary representative used to construct these queries has norm at most $R$, and $\sigma_{\min}(X_i) \geq \kappa$ for all $i \in J$ at every sampling call. The norm bound includes auxiliary
representatives for facets outside $J$. Assume that $M(\Delta) \to +\infty$ as $\Delta \downarrow 0$ and $\Delta(M(\Delta)+K-1)<\eta_S$ for all sufficiently small $\Delta$. Then, every run terminates after at most $\left(1+\frac{2RL_e}{\Delta}\right)^{nd}$ inverse optimization updates and its normalized output satisfies $P^a_{M(\Delta),\Delta} \rightarrow \frac{P^{DM}}{\|P^{DM}\|_F}$. As a consequence, we have $L_\textnormal{pred}(P^a_{M(\Delta),\Delta},P^{DM},x) \rightarrow 0$ for every $x \notin \ker(P^{DM})$. 
\end{lemma}

\subsection{Experiment setup}\label{sec:app:exp-setting}
\paragraph{Synthetic instances.} In the experiment, the decision-maker first learns the prediction model $P^{DM}$ by performing a linear regression on a randomly generated dataset $D=\{x_i,c_i\}$, where all coordinates of $x_i$ are independently drawn from $U[0,20]$. The responses satisfy $c_i=P^\textnormal{true}x_i+\varepsilon$, where $\varepsilon \sim \mathcal{N}(0,I_n)$. Here, $P^\textnormal{true}$ is a randomly generated ground-truth prediction matrix, unknown to the decision-maker, with $P^\textnormal{true}_{ij} \sim \mathcal{U}[-10,10]$. Given a query context $x$, the decision-maker will predict the cost vector $P^{DM}x$ and solve the downstream linear optimization problem $\min_{w \in S} (P^{DM}x)^\top w$. The downstream optimization problem is a fractional Knapsack problem:
\begin{equation} \label{eq:knapsack-prob}
\begin{array}{cl} 
\min_{w \in \R^n} & (P^{DM}x)^\top w \\ 
\textnormal{s.t. } & a^\top w \leq b, \ \textbf{0} \leq w \leq \mathbf{1}. 
\end{array}
\end{equation}
where $a \in \R^n$ is a vector whose entries start at $1$ and increase by $0.5$ successively and $b=0.6\sum_{i=1}^na_i$. We use the fractional Knapsack problem because it models resource allocation decisions, including those arising in recommendation systems. The number of vertices can grow exponentially with $n$, which affects the size of a complete boundary sampling call. The optimization problem in Eq.~\eqref{eq:subopt-reform}, however, is a linear program using the inequality representation of the polytope and does not require vertex enumeration.

We compare boundary sampling (BndS), uniform sampling (Unif), and a nonadaptive inverse optimization baseline (INV). Unif and BndS alternate sampling and inverse optimization as in Algorithm~\ref{algo:attack}, using the sampling constructions in Algorithms~\ref{algo:unif-sampling} and~\ref{algo:edge-sampling}, respectively. INV performs one inverse optimization fit on a nonadaptive batch of contexts, with sample sizes matched in the reported comparisons. Unif draws predicted cost directions uniformly from the unit sphere in $\R^n$ and maps them to contexts using the current attacker matrix. The INV batch is selected without adapting to an updated attacker matrix. Unless otherwise specified, all three methods use the suboptimality-loss formulation in Eq.~\eqref{eq:subopt-reform}.

For the attack comparisons, we perform 100 repetitions, using a different deployed matrix $P^{DM}$ in each repetition. We report the median prediction loss and the interval between its 25th and 75th percentiles.

In the query-budget experiment, we use $d=10$ and $n=5$, with 100 queries per iteration. MIX allocates 50 queries to boundary sampling and 50 to uniform sampling. Each parameter setting is repeated five times, and we report the median prediction loss. Here, BndS and the boundary component of MIX use a budgeted variant of Algorithm~\ref{algo:edge-sampling} rather than its complete query set, while Theorem~\ref{thm:sample-global-error} is proved for the complete construction.

For each defensive parameter setting, we perform 100 repetitions and record the attacker's prediction loss and the decision-maker's regret. The parameter is $\epsilon$ for decision perturbation and $\sigma$ for prediction perturbation. Each point in a trade-off plot reports the median regret and median prediction
loss over the repetitions, with the 25th and 75th percentiles of prediction loss shown as an interval.

All experiments were conducted on a machine with two AMD EPYC 7643 48-core processors, with each job allocated one CPU core and 16 GB of memory.

\paragraph{Real data.} The KuaiRand-1K dataset~\citep{Gao-2022-Kuairand} contains $1000$ users and approximately four million videos. Its standard policy portion contains $11713045$ interactions. We use the records of $800$ users for training and the remaining users for testing. Each user is represented by a feature vector of dimension $d=18$. We cluster the videos into $n=5$ types and then estimate each user’s click probability for every video type from the interaction logs. Each log entry records the interaction between a specific user and a specific video. Using the cluster label of the video and the observed interaction, we update the user’s click-through rate for the corresponding video type. The platform fits a linear predictor of click-through rates and defines $P^{DM}$ as the negative of its coefficient matrix. Thus, the predicted click-through-rate vector is $-P^{DM}x$. With $k=\lceil \frac{n}{2}\rceil$, the recommendation is obtained from
\begin{equation}
\begin{array}{cl}
\min_{w \in \R^n} & (P^{DM}x)^\top w \\
\textnormal{s.t.} & \sum_{i=1}^n w_i \leq k,\ \textbf{0} \leq w \leq \mathbf{1}. 
\end{array}
\end{equation}
Prediction loss is evaluated using the attacker and deployed cost vectors, and regret is measured under the fitted cost vector $P^{DM}x$ rather than realized clicks.
\begin{figure}[!t]
\centering
\includegraphics[width=0.48\textwidth]{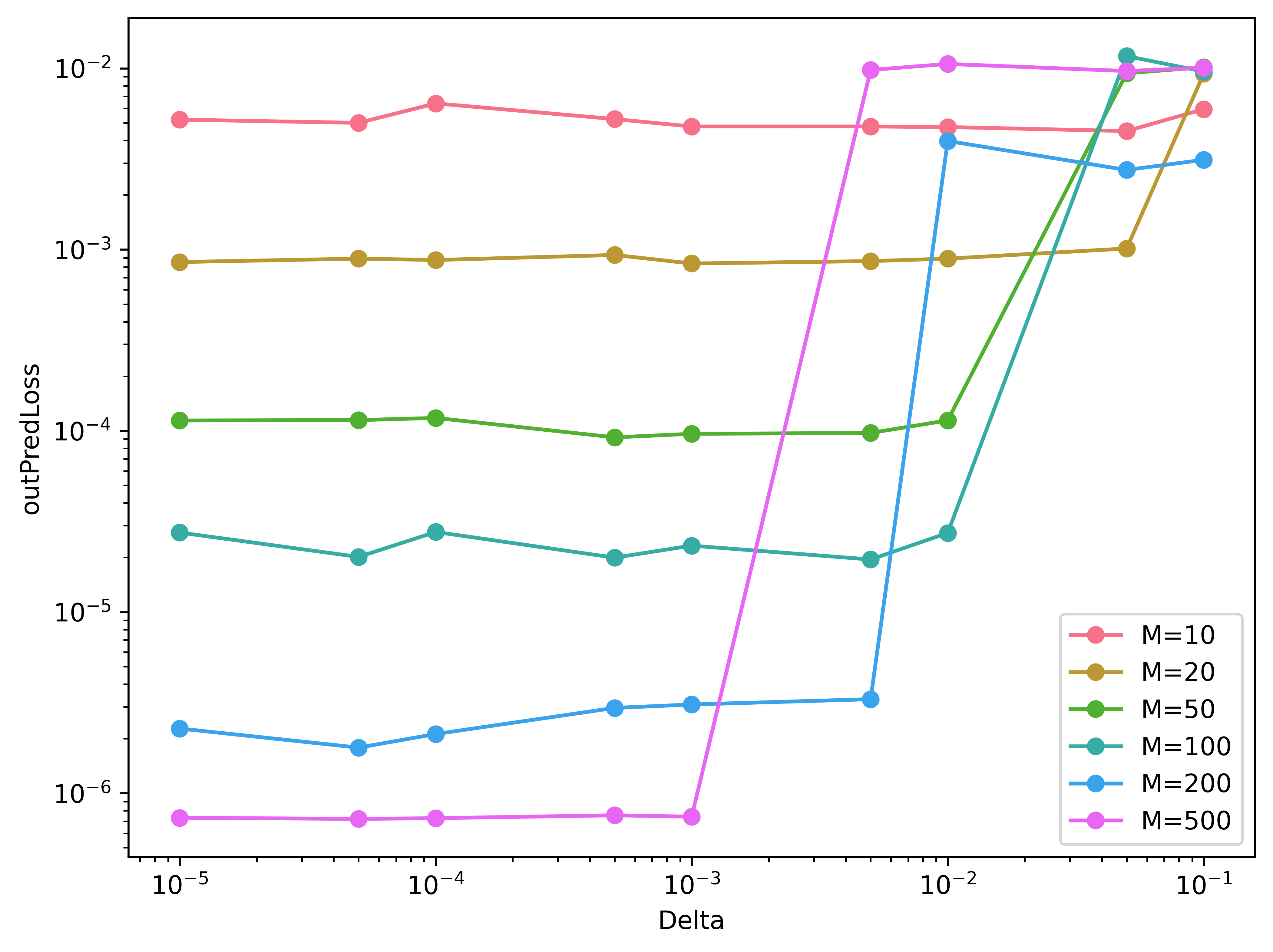} \label{subfig:ADMM-converge}
\includegraphics[width=0.48\textwidth]{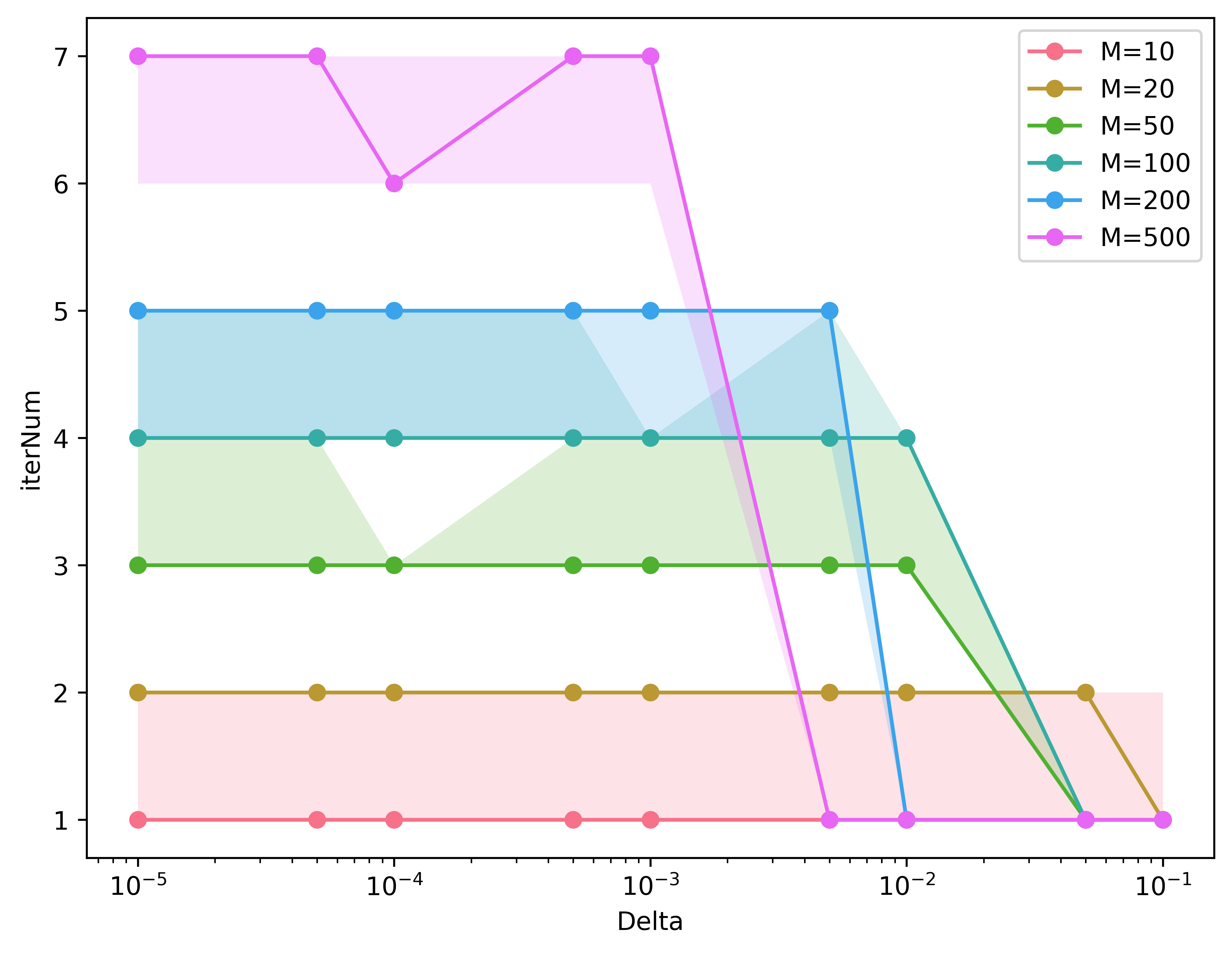} \label{subfig:Heur-converge}
\caption{The out-of-sample prediction loss and iterations for the attacking framework in different attacker's parameter settings.} \label{fig:coefCompare}
\end{figure}
\begin{figure*}[!t]
\centering
\begin{subfigure}[t]{0.32\textwidth}
\centering
\includegraphics[width=\textwidth]{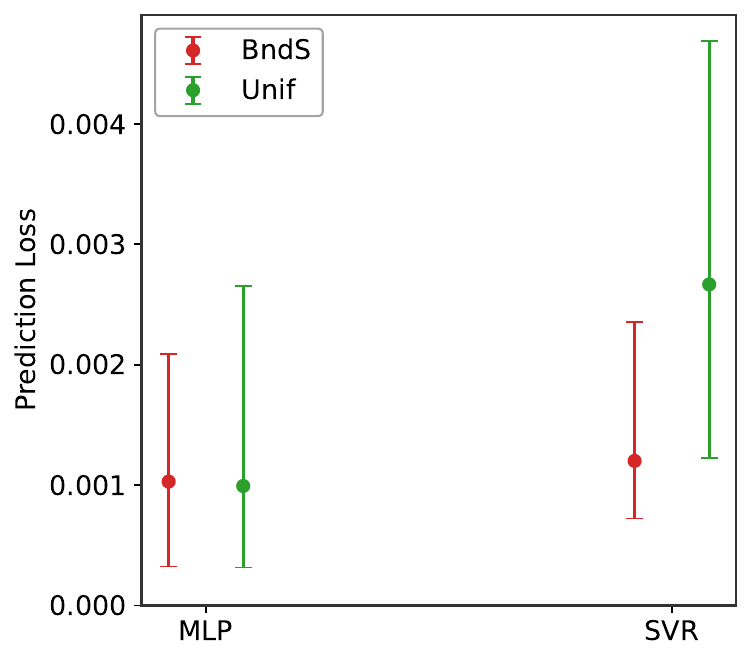}
\caption{}
\label{subfig:misspec-predloss}
\end{subfigure}
\begin{subfigure}[t]{0.32\textwidth}
\centering
\includegraphics[width=\textwidth]{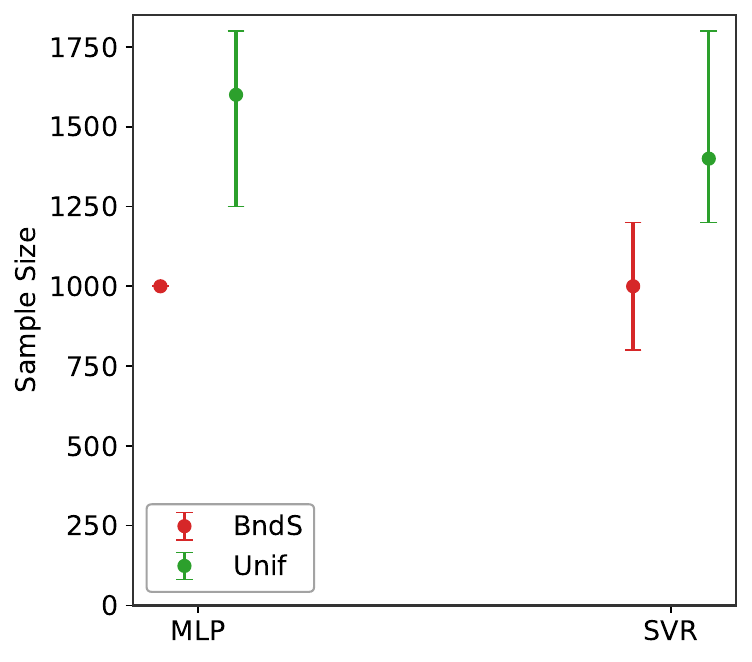}
\caption{}
\label{subfig:misspec-samplesize}
\end{subfigure}
\begin{subfigure}[t]{0.32\textwidth}
\centering
\includegraphics[width=\textwidth]{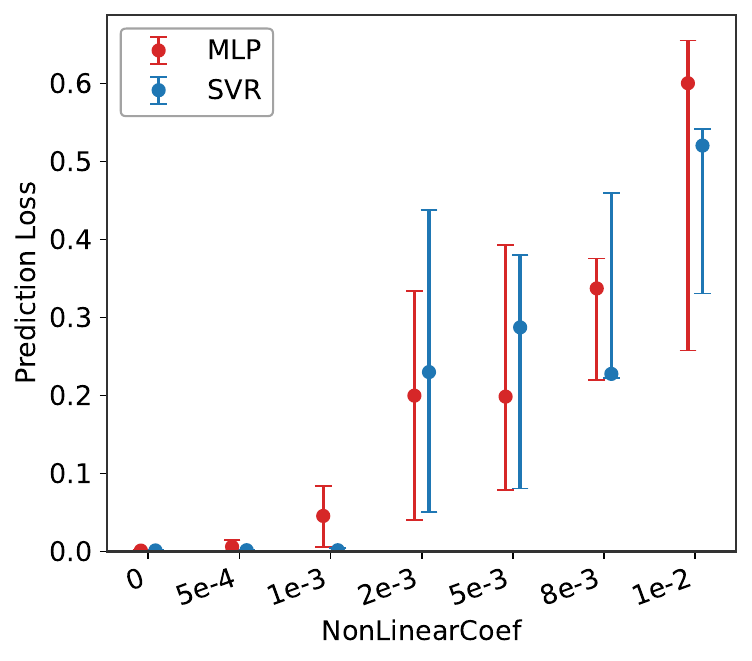}
\caption{}
\label{subfig:misspec-nonlinear}
\end{subfigure}
\caption{(a) Prediction loss; (b) Sample size for BndS and Unif against MLP and SVR predictors fitted to the linear synthetic data; and (c) Prediction loss of a linear attacker against MLP and SVR predictors as the quadratic coefficient $\gamma$ varies.}
\label{fig:misspec}
\end{figure*}

\subsection{Additional experimental results} \label{sec:app:exp}
\paragraph{Attacker's coefficient.} Figure~\ref{fig:coefCompare} compares prediction loss and the number of iterations for different values of $M$ and $\Delta$. In the tested settings, larger $\Delta$ is generally associated with fewer iterations and greater prediction loss. For fixed $M$, both quantities change little once $\Delta$ is sufficiently small. The condition $\Delta(M+K-1)<\eta_S$ is sufficient for the stopping argument in Theorem~\ref{thm:sample-global-error} but the figure does not establish that it is a sharp threshold.

Proposition~\ref{prop:adv-algo-stop} bounds the number of inverse optimization updates rather than specifying its realized value. With a common query-norm bound $R$, the bound has no explicit dependence on $M$, although the observed number of iterations can depend on $M$. Prediction recovery in Theorem~\ref{thm:sample-global-error} concerns a coupled limit in $M$ and $\Delta$ under uniform sampling regularity.
\begin{figure}[!t]
\centering
\begin{subfigure}[t]{0.48\textwidth}
\centering
\includegraphics[width=\textwidth]{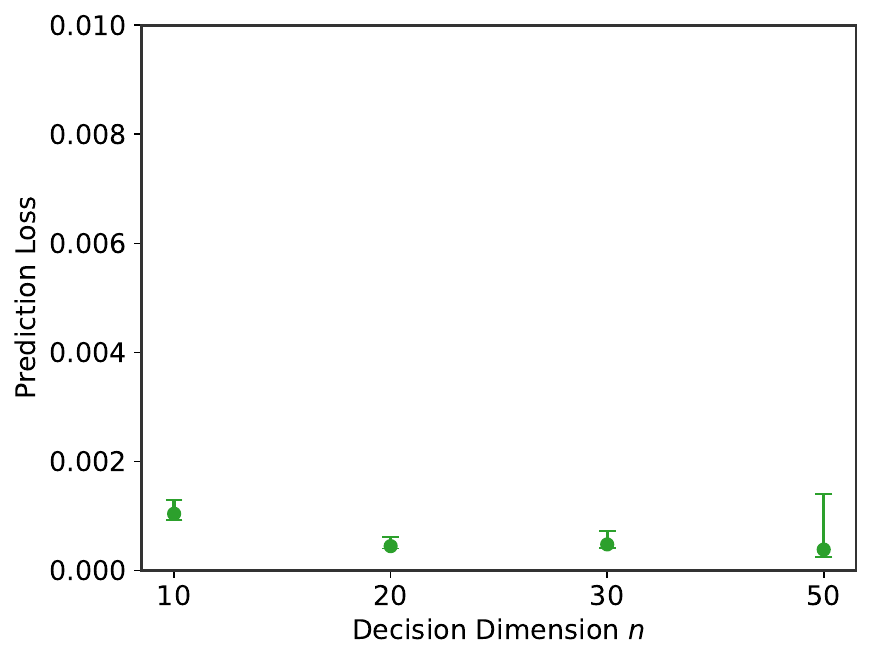}
\caption{Prediction Loss}
\label{subfig:high-dim-attacker-loss}
\end{subfigure}
\hfill
\begin{subfigure}[t]{0.48\textwidth}
\centering
\includegraphics[width=\textwidth]{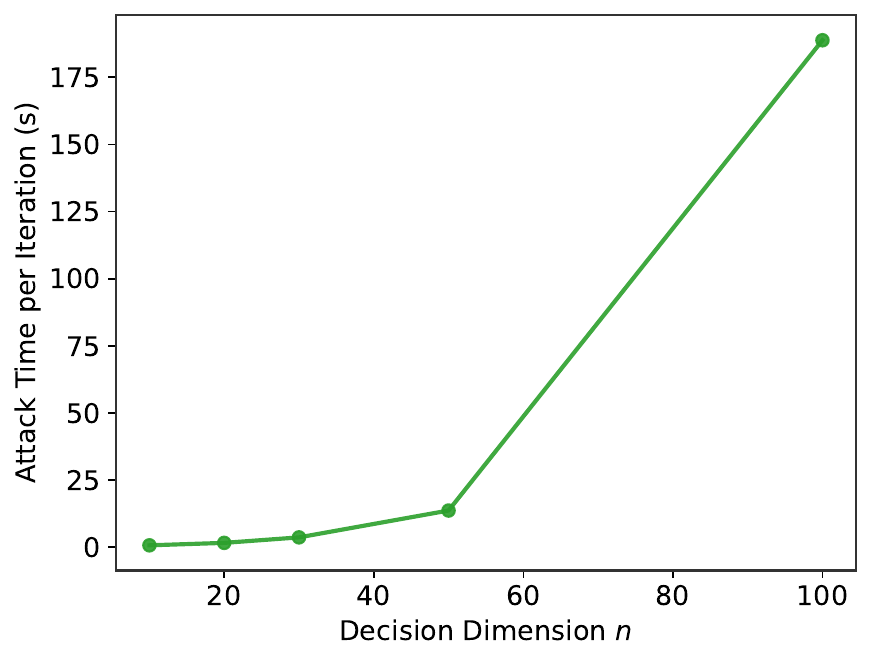}
\caption{Time}
\label{subfig:high-dim-attacker-time}
\end{subfigure}
\caption{Prediction loss (a) and runtime per iteration (b) of Unif on fractional Knapsack instances of different dimensions.}
\label{fig:high-dim-attacker}
\end{figure}
\begin{figure}[!t]
\centering
\includegraphics[width=0.5\linewidth]{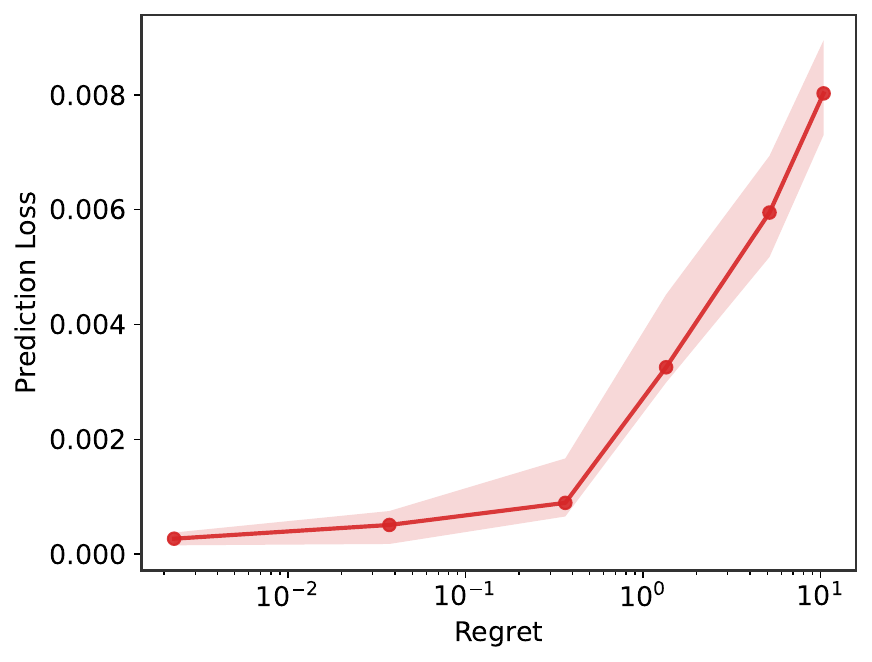}
\caption{Security-performance trade-off under the fractional Knapsack problem with $n=20$, using at most three neighboring vertices per query.} \label{fig:high-dim-DM}
\end{figure}
\begin{figure}[!t]
\centering
\includegraphics[width=0.48\textwidth]{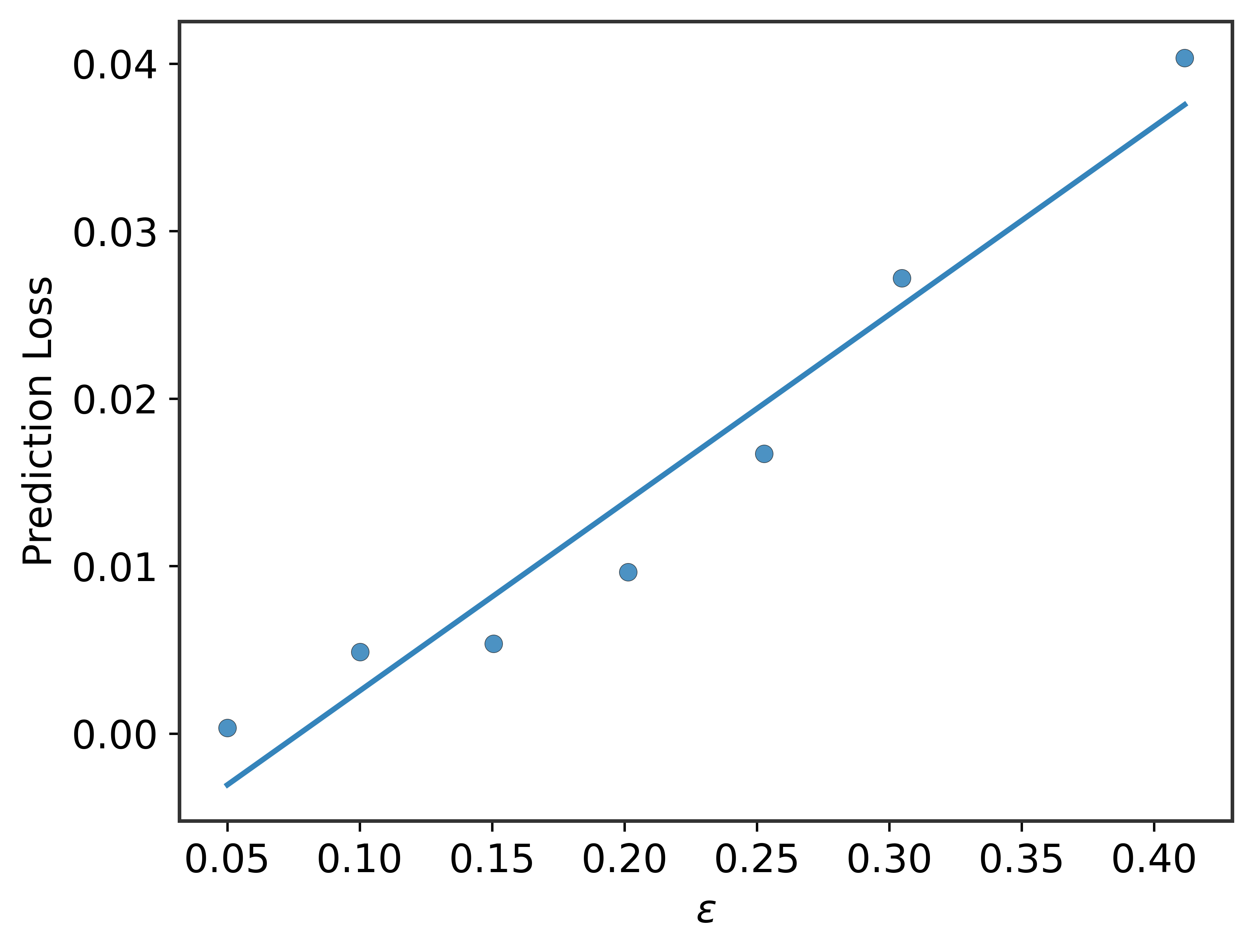} \label{subfig:DM-eps-predLoss}
\includegraphics[width=0.48\textwidth]{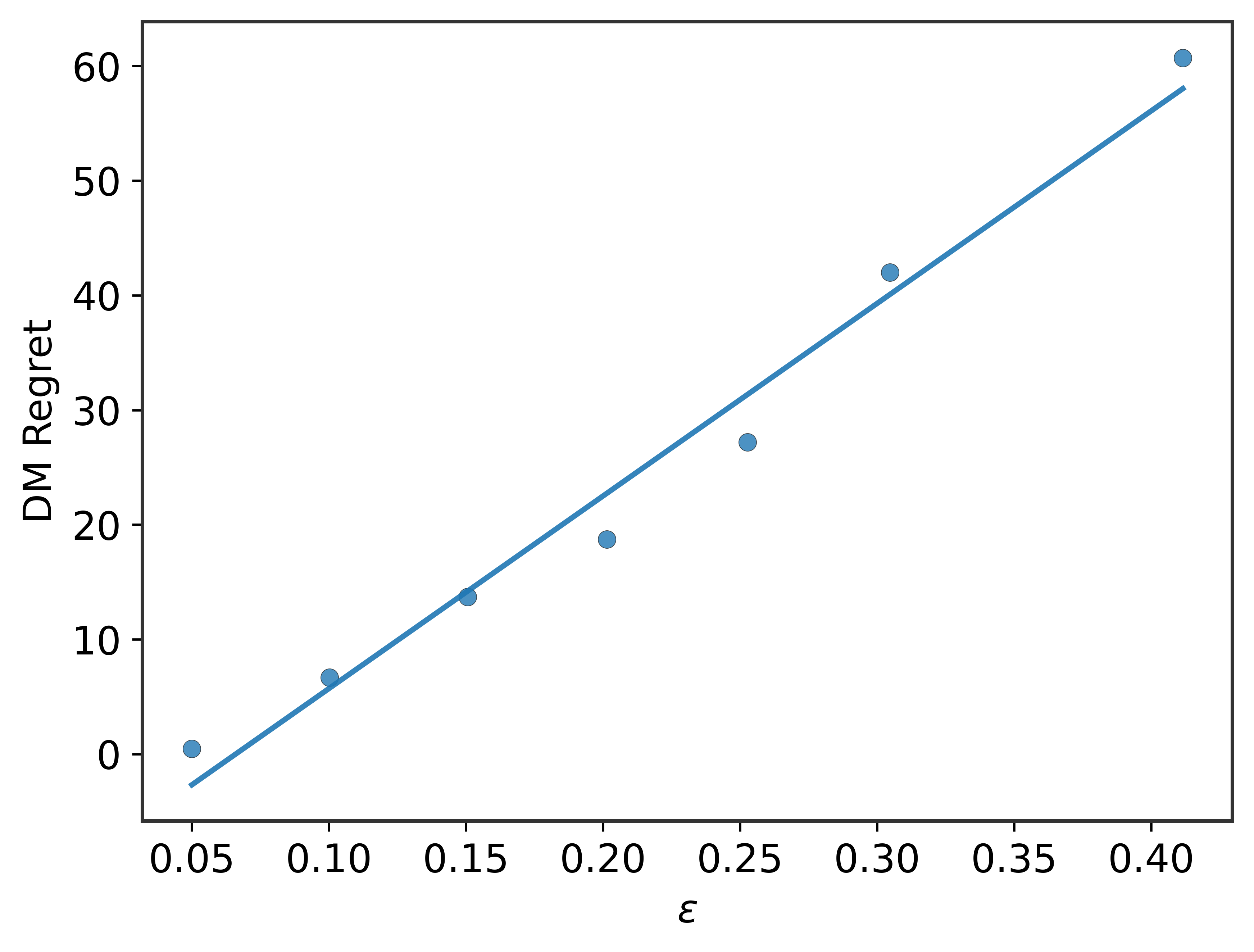} \label{subfig:DM-eps-regret}
\caption{Prediction loss (left) and decision-maker's regret (right) as functions of the defensive parameter $\epsilon$.} \label{fig:DM-eps}
\end{figure}

\paragraph{Attacker's robustness to misspecification.} On the synthetic dataset, we evaluate the linear attack against two nonlinear deployed model classes: a multilayer perceptron (MLP) with one hidden layer and ReLU activation, and support vector regression (SVR) with a Gaussian kernel. The hyperparameters,
including hidden-layer width and kernel bandwidth, are selected by cross-validation.

Figure~\ref{subfig:misspec-predloss} reports prediction-loss values below $0.01$ for the evaluated linear attacks against both deployed models, while Figure~\ref{subfig:misspec-samplesize} reports their sample sizes. These results show that a linear attacker can approximate the deployed prediction directions in these experiments even when the decision-maker uses a nonlinear model class.

We also evaluate the linear attacker on data with a quadratic component:
\begin{equation*}
c=P^\textnormal{true}x+\gamma B\phi_2(x)+\epsilon.
\end{equation*}
Here, $\phi_2(x)$ collects all quadratic terms of $x$, $P^\textnormal{true}_{ij} \sim \mathcal{U}[-10,10]$,
$B_{ij} \sim \operatorname{Bernoulli}(0.5)$, and $\epsilon$ is a zero-mean Gaussian noise vector. The decision-maker fits the MLP or SVR model to these data, while the attacker retains its linear model class. Figure~\ref{subfig:misspec-nonlinear} shows generally larger prediction loss for larger values of $\gamma$ over the tested range.

\paragraph{Large-scale optimization problem.} A call to Algorithm~\ref{algo:edge-sampling} makes $(d-n+1)(m+\sum_{k=1}^{|V|} |I_k|)$ query calls, counting repeated contexts separately. This count can grow exponentially with $n$ through the number of vertices. Context construction, oracle evaluations, and inverse optimization also contribute to the computational cost. The cost of Algorithm~\ref{algo:DM-strategy} likewise depends on how neighboring vertices and their angular eligibility are computed. We evaluate uniform sampling and a restricted-neighbor defense on larger fractional Knapsack instances.

To control the number of queries in larger problems, we use uniform sampling on the fractional Knapsack problem in Eq.~\eqref{eq:knapsack-prob}, with $d=2n$ and $N=5n$ new queries per iteration. The number of new queries is therefore linear in $n$, while the size of the inverse-fitting problem depends on the accumulated dataset. This choice does not by itself establish a polynomial bound on total attack time. Figure~\ref{subfig:high-dim-attacker-loss} reports small prediction losses without defense, and Figure~\ref{subfig:high-dim-attacker-time} reports increasing time per iteration as $n$ grows.

For the defender, we evaluate decision perturbation restricted to at most three neighboring vertices per query against a uniform-sampling attacker on fractional Knapsack instances with $n=20$. Figure~\ref{fig:high-dim-DM} shows a security-performance trade-off for this implementation. The regret bound in Proposition~\ref{prop:DM-regret} remains valid provided every selected neighbor passes the angular
eligibility test in Algorithm~\ref{algo:DM-strategy}. The expected prediction error bound in Theorem~\ref{thm:DM-def-BS} assumes that Algorithm \ref{algo:DM-strategy} includes every neighboring vertex satisfying the angular eligibility condition..

\paragraph{Prediction loss and regret on defensive parameter.} Figure~\ref{fig:DM-eps} shows that prediction loss and regret generally increase over the tested values of $\epsilon$. This fits our theoretical result. Proposition~\ref{prop:DM-regret} gives a regret upper bound linear in $\epsilon$, while Theorem~\ref{thm:DM-def-BS} gives a nondecreasing lower bound on expected prediction loss under its population-fitting condition. 

\section{Proof}\label{sec:app:proof}
\begin{proof}[Proof of Proposition \ref{prop:generalize-classification}]
Suppose that $s(x)=-p(x)$ is the class utility vector. Then, the classifier selects a label by finding $i(x) \in \argmin_{1 \leq i \leq n} p_i(x)$. For every $w \in S$, we have
\begin{equation*}
w^\top p(x) = \sum_{i=1}^n w_i p_i(x) \geq \min_i p_i(x),
\end{equation*}
with equality at $e_{i(x)}$ exactly when $i(x) \in \argmin_{1 \leq i \leq n} p_i(x)$. Matching the tie-breaking rules therefore identifies each returned class label $i(x)$ with the returned vertex $e_{i(x)}$, and conversely.

\end{proof}

\begin{proof}[Proof of Lemma \ref{lem:invariant-calibration}]
The key is to show that calibration is independent of the particular inverse feasibility loss. Indeed, we define the set of globally decision consistent matrices by
\begin{equation*} 
\mathcal{F} = \{P \in \R^{n \times d} \setminus \{0\}: w^\star(P^{DM}x) \in W^\star(Px) \textnormal{ for every } x \notin \ker(P^{DM})\},
\end{equation*}

and the set of matrices with globally correct prediction directions by 
\begin{equation*} 
\mathcal{G} = \{P \in \R^{n \times d} \setminus \{0\}: L_\textnormal{pred}(P,P^{DM};x)=0 \textnormal{ for every } x \notin \ker(P^{DM})\}.
\end{equation*}
For any inverse feasibility loss $L$, we obtain that $L(P,P^{DM};x)=0$ if and only if $w^\star(P^{DM}x) \in W^\star(Px)$ for every nonzero matrix $P$ and every context $x$. This implies that all inverse feasibility losses have the same global zero loss set $\mathcal{F}$. Then, we establish the relationship between $\mathcal{F}$ and $\mathcal{G}$. 

We claim that $\mathcal{G} \subseteq \mathcal{F}$. Indeed, we take any $P \in \mathcal{G}$ and $x \notin \ker(P^{DM})$. Since $P^{DM}x \neq 0$ and $L_\textnormal{pred}(P,P^{DM};x)=0$, we have $Px=\gamma_x P^{DM}x$ for some $\gamma_x>0$. This implies $W^\star(Px)=W^\star(P^{DM}x)$ and $w^\star(P^{DM}x) \in W^\star(Px)$. Since this holds at every nonkernel context, $P \in \mathcal{F}$. This yields the desired claim. 

It suffices to consider two cases. If $\mathcal{F}=\mathcal{G}$, the calibration holds for every inverse feasibility loss. Otherwise, there exists a matrix $P_\textnormal{bad} \in \mathcal{F} \setminus \mathcal{G}$. By definition, we have (i) $L(P_\textnormal{bad},P^{DM};x)=0$ for every $x \notin \ker(P^{DM})$ and (ii) there exists a context $x_0 \notin \ker(P^{DM})$ such that $L_\textnormal{pred}(P_\textnormal{bad},P^{DM};x_0)>0$, for every inverse feasibility loss $L$. In this case, none of the inverse
feasibility losses calibrates the prediction loss. 

We now show that calibration fails for an $n$-dimensional hypercube. Indeed, for every $w \in S$, we have $w=w_0+\ell Uz$ where $w_0 \in \R^n$, $\ell>0$, $z \in [0, 1]^n$ and $U$ is an orthogonal matrix.  This implies that minimizing $c^\top w$ over $S$ is equivalent to minimizing $(U^\top c)^\top z$ over $[0,1]^n$. This problem separates coordinate-wise: an optimal $z^\star$ satisfies
\begin{equation*}
z_i^\star \in \begin{cases} \{0\}, & \textnormal{if } (U^\top c)_i > 0,\\ \{1\},  & \textnormal{if } (U^\top c)_i < 0,\\ [0,1], & \textnormal{otherwise}. \end{cases} \textnormal{ for all } i=1, 2, \ldots, n.  
\end{equation*}
We define $D=\operatorname{diag}(1,\ldots,1,2)$ and $P^a=UDU^\top P^{DM}$. Since $UDU^\top$ is invertible and $P^{DM}$ has full row rank, $P^a$ has full row rank and is an admissible nonzero attacker matrix. For every context $x$, we have $U^\top P^a x=DU^\top P^{DM}x$. Since all diagonal entries of $D$ are positive, these transformed cost vectors have identical coordinate-wise signs. The preceding characterization of the optimal solution set therefore gives $W^\star(P^a x)=W^\star(P^{DM}x)$ for every $x \in \R^d$. In particular, every decision returned by the deployed model is optimal under the attacker cost. Thus, every inverse feasibility loss satisfies $L(P^a,P^{DM};x)=0$ for every $x \in \R^d$.

Since $P^{DM}$ has full row rank and $n \geq 2$, there exists $x_0 \in \R^d$ such that $P^{DM}x_0 = U(e_1+e_n)$. By definition, we have $P^a x_0=U(e_1+2e_n)$. Putting these pieces together yields 
\begin{equation*}
L_\textnormal{pred}(P^a,P^{DM};x_0) = \angle(e_1+2e_n,e_1+e_n) = \arccos\left(\tfrac{3}{\sqrt{10}}\right) > 0.
\end{equation*}
This implies that $x_0\notin\ker(P^{DM})$ and $P^a \in \mathcal{F} \setminus \mathcal{G}$. Therefore, we conclude that no inverse feasibility loss calibrates the prediction loss on an $n$-dimensional hypercube.
\end{proof}

For the remaining proofs, we let
\begin{equation*}
r_i = -\tfrac{a_i}{\|a_i\|},\quad C_k = \cone(w_k),\quad I_k=\{i:a_i^\top w_k=b_i\},\quad \mathcal{C}_i = \{k:i \in I_k\}, 
\end{equation*}
where $I_k$ indexes the facets containing $w_k$, $\mathcal{C}_i$ indexes the vertices of facet $i$, and $C_k$ is closed, full-dimensional and pointed. In our analysis, we will use three facts: (i) $c \in \operatorname{int}(C_k) \Leftrightarrow W^\star(c)=\{w_k\}$, (ii) $C_k=\operatorname{polycone}\{r_i:i\in I_k\}$ and (iii) $\cap_{k \in \mathcal{C}_i} C_k = \{\alpha r_i: \alpha \geq 0\}$. The last fact follows since a cost vector makes every point of facet $i$ optimal if and only if it is a nonnegative multiple of the inward normal $r_i$.

\begin{proof}[Proof of Theorem \ref{thm:loss-cali-cond}]
Suppose that $P^a \neq 0$ satisfies $w^\star(P^{DM}x) \in W^\star(P^a x)$ for every $x \notin \ker(P^{DM})$. Then, we show that $\ker(P^{DM}) \subseteq \ker(P^a)$. Indeed, we fix $z \in \ker(P^{DM})$, choose $c_0 \in \operatorname{int}(C_k)$ for some vertex $w_k$, and take $x_0$ satisfying $P^{DM}x_0=c_0$. Since $P^{DM}$ has full row rank, such $x_0$ exists. Then, we have $P^{DM}(x_0+tz) = c_0 \neq 0$ for every $t \in \R$ and hence $W^\star(c_0)=\{w_k\}$. The consistency condition implies $P^a(x_0+tz) \in C_k$ for every $t \in \R$. Since $C_k$ is a closed cone, $P^a z,-P^a z\in C_k$. Since $C_k$ is pointed, $P^a z=0$. Thus, we have $\ker(P^{DM}) \subseteq \ker(P^a)$. 

Since $P^{DM}$ has full row rank and $\ker(P^{DM}) \subseteq \ker(P^a)$, we can write $P^a=BP^{DM}$ for a linear map $B: \R^n \to \R^n$. It suffices to show that $B=\gamma I_n$ for some $\gamma>0$. Indeed, for any $c \in \operatorname{int}(C_k)$, we choose $x$ with $P^{DM}x=c$. Consistency gives $w_k \in  W^\star(Bc)$, so $Bc \in C_k$. Since $C_k = \overline{\operatorname{int}(C_k)}$, continuity gives $BC_k \subseteq C_k$ for every $k$. This implies that, for each facet $i$, we have $Br_i \in \bigcap_{k \in \mathcal{C}_i} C_k = \{\alpha r_i: \alpha \geq 0\}$. Thus, we have $Ba_i = \lambda_i a_i$ for some $\lambda_i \geq 0$. Applying $B$ to both sides of $a_{k_{n+1}}=\sum_{j=1}^n\beta_j a_{k_j}$ in Assumption~\ref{assumption:main} yields $\sum_{j=1}^n \beta_j(\lambda_{k_j}-\lambda_{k_{n+1}})a_{k_j} = 0$. Linear independence and $\beta_j \neq 0$ give $\lambda_{k_j}=\lambda_{k_{n+1}}$ for every $j=1,\ldots,n$. Since $a_{k_1},\ldots,a_{k_n}$ form a basis of $\R^n$, we have $B=\gamma I_n$, where $\gamma=\lambda_{k_{n+1}} \geq 0$. As $P^a \neq 0$, we have $\gamma>0$ and $P^a=\gamma P^{DM}$. 
\end{proof}

\begin{proof}[Proof of Proposition \ref{prop:edge-sample}]
The proof of Theorem~\ref{thm:loss-cali-cond} shows that global decision consistency alone gives a linear map $B: \R^n \to \R^n$ satisfying $P^a=BP^{DM}$, $Br_i=\lambda_i r_i$ and $\lambda_i \geq 0$ for every $i$. Since $\operatorname{rank}(P^a)=n$, the relation $P^a=BP^{DM}$ implies that $B$ is invertible. Since $r_i \neq 0$, we have $\lambda_i>0$ for every $i$.

Suppose that $\bar{x}$ is the boundary context, so that $P^a\bar{x}=\alpha r_i$ for some facet $i$ and $\alpha>0$. Then, by using $B^{-1}r_i=\lambda_i^{-1}r_i$, we have
\begin{equation*}
P^{DM}\bar{x} = B^{-1}P^a\bar{x} = \tfrac{\alpha}{\lambda_i}r_i = \tfrac{1}{\lambda_i}P^a\bar{x}.
\end{equation*}
This implies that the two costs are nonzero and positively aligned. Thus, we have $L_\textnormal{pred}(P^a,P^{DM};\bar{x})=0$. 
\end{proof}

\begin{proof}[Proof of Proposition \ref{prop:adv-algo-stop}]
Since $\|P^0\|_F=1$ and every update is normalized, we have $\|P^t\|_F=1$ for every $t$. By the zero-loss assumption and invariance of inverse feasibility under positive scaling, each $P^s$, $s\geq1$, is inverse feasible on $D^{s-1}$. At any nonterminal iteration $t$, there exist $(x_t,w_t) \in D^t$ and $v_t \in V$ such that
\begin{equation*}
(P^t x_t)^\top(w_t-v_t) > \Delta.
\end{equation*}
For every subsequent iterate $P^{t'}$, we have $D^t \subseteq D^{t'-1}$. This implies $(P^{t'} x_t)^\top(w_t-v_t) \leq 0$. Thus, we have
\begin{equation*}
\Delta < ((P^t-P^{t'})x_t)^\top(w_t-v_t) \leq \|P^t-P^{t'}\|_F\|x_t\|\|w_t-v_t\| \leq RL_e\|P^t-P^{t'}\|_F.
\end{equation*}
Thus, any two nonterminal iterates are separated by more than $\frac{\Delta}{RL_e}$ in Frobenius norm.

For any finite collection of $T$ nonterminal iterates, the open Frobenius balls of radius $\frac{\Delta}{2R L_e}$ centered at these iterates are disjoint. Since every iterate has unit Frobenius norm, these balls lie in the ball centered at zero with radius $1+\frac{\Delta}{2RL_e}$. Then, we have
\begin{equation*}
T\left(\tfrac{\Delta}{2RL_e}\right)^{nd} \leq \left(1+\tfrac{\Delta}{2RL_e}\right)^{nd},
\end{equation*}
which implies
\begin{equation*}
T \leq \left(1+\tfrac{2RL_e}{\Delta}\right)^{nd}.
\end{equation*}
Each nonterminal iteration performs one inverse optimization update, and each sampling call is finite.
Therefore, Algorithm~\ref{algo:attack} terminates after at most $(1+\frac{2RL_e}{\Delta})^{nd}$ inverse optimization updates. 
\end{proof}

\begin{lemma}\label{lem:boundary-provide-acc-info}
Let $P^a$ have full row rank, and let $\bar{x} \in \R^d$ satisfy $P^a\bar{x}=\alpha r_i$ for some facet $i$ and $\alpha>0$. Suppose that, for some $\epsilon>0$, we have $w^\star(P^{DM}x) \in W^\star(P^a x)$ for every $x \in \R^d$ such that $P^{DM}x \neq 0$, $P^a x \neq 0$ and $\angle(P^a x,r_i) < \epsilon$. Then, we have 
\begin{equation*}
P^{DM}\bar{x} \in \{\alpha r_i: \alpha \geq 0\}.
\end{equation*}
In particular, if $P^{DM}\bar{x} \neq 0$, we have $L_\textnormal{pred}(P^a,P^{DM};\bar{x})=0$.
\end{lemma}
\begin{proof}
If $P^{DM}\bar{x}=0$, the conclusion is immediate. Otherwise, we fix $k \in \mathcal{C}_i$. Since $P^a$ has full row rank, we choose $h_k \in \R^d$ with $P^a h_k \in \operatorname{int}(C_k)$ and define $x_t = \bar{x}+th_k$ for any $t>0$. Then, we have $x_t \to \bar{x}$ as $t \downarrow 0$ and
\begin{equation*}
P^a x_t = \alpha r_i + tP^a h_k \in \operatorname{int}(C_k).
\end{equation*}
For sufficiently small $t>0$, we have $P^{DM}x_t \neq 0$ and $\angle(P^a x_t,r_i) < \epsilon$. Consistency guarantees $w^\star(P^{DM}x_t) \in W^\star(P^a x_t)=\{w_k\}$ so we have $P^{DM}x_t \in C_k$. Closedness guarantees $P^{DM}\bar{x} \in C_k$ which holds for every $k \in \mathcal{C}_i$. Thus, we have
\begin{equation*}
P^{DM}\bar{x} \in \bigcap_{k \in \mathcal{C}_i} C_k = \{\alpha r_i: \alpha \geq 0\}.
\end{equation*}
If $P^{DM}\bar x\neq0$, this implies that $P^{DM}\bar{x}$ is positively aligned with $P^a\bar{x}=\alpha r_i$ and $L_\textnormal{pred}(P^a,P^{DM};\bar{x})=0$.  
\end{proof}

\begin{proof}[Proof of Proposition \ref{prop:info-near-boundary}]
Suppose that no $\epsilon$-near-boundary context belongs to $I(P^a,P^{DM})$. Then, by using the definition of $I(P^a,P^{DM})$, every such context has zero suboptimality loss. Thus, we have $w^\star(P^{DM}x) \in W^\star(P^a x)$ whenever $P^a x \neq 0$ and $\angle(P^a x,r_i)<\epsilon$ for some facet $i$. By Lemma~\ref{lem:boundary-provide-acc-info}, we have
\begin{equation*}
P^{DM}\bar{x} \in \{\alpha r_i: \alpha \geq 0\} \textnormal{ whenever } P^a\bar{x}=r_i.
\end{equation*}
This implies that $P^{DM} \in \mathcal{P}_c(P^a)$. By Assumption~\ref{assumption:main} and
Proposition~\ref{prop:1-dim-suff-condition}, we have $P^{DM}=\gamma P^a$ for some $\gamma \in \R$.
The displayed ray membership and $P^{DM} \neq 0$ imply $\gamma>0$. This implies that the prediction loss is zero at every context, yielding a contradiction. Therefore, an $\epsilon$-near-boundary context in $I(P^a,P^{DM})$ must exist.
\end{proof}

\begin{proof}[Proof of Theorem 3.9]
Theorem~\ref{thm:sample-global-error} shows finite termination and $P^a_{M(\Delta),\Delta} \rightarrow \frac{P^{DM}}{\|P^{DM}\|_F}$. For every $x \notin \ker(P^{DM})$, the limiting cost is nonzero. Therefore, the continuity of angular distance gives the desired result. 
\end{proof}

\begin{proof}[Proof of Proposition \ref{prop:DM-regret}]
Fix a cache realization and a context $x$, and write $c=P^{DM}x$, $w_0=w^\star(c)$ and $w=\tilde{w}(x)$. If $c=0$ or $w=w_0$, the regret is zero. Otherwise, Algorithm~\ref{algo:DM-strategy} gives $\angle(c,\cone(w)) \leq \epsilon$. The intersection of $\cone(w)$ with the unit sphere is nonempty and compact. We choose a unit vector $u \in \cone(w)$ satisfying
\begin{equation*}
\angle(c,u) = \angle(c,\cone(w)) \leq \epsilon.
\end{equation*}
Since $w$ is optimal under $u$, we have $u^\top(w-w_0) \leq 0$. Thus, we have
\begin{equation*}
c^\top(w-w_0) \leq \|c\|\left(\tfrac{c}{\|c\|}-u\right)^\top(w-w_0) \leq \|c\|\|w-w_0\|\left\|\tfrac{c}{\|c\|}-u\right\| \leq 2L_e\|c\|\sin\left(\tfrac{\angle(c,u)}{2}\right) \leq L_e\epsilon\|c\|.
\end{equation*} 
Taking expectation under $x\sim Q$ yields
\begin{equation*}
\mathbb{E}_{x\sim Q}[\textnormal{Regret}(\tilde{w}(x);x)] \leq L_e\epsilon\mathbb{E}_{x\sim Q}[\|P^{DM}x\|]. 
\end{equation*}
This completes the proof. 
\end{proof}

\begin{proof}[Proof of Theorem \ref{thm:DM-def-BS}]
Fix a cache realization and a context $x$ with nonzero costs, and write $c=P^{DM}x$, $c_a=P^a x$, $w_0=w^\star(c)$ and $w_a=w^\star(c_a)$. The boundary of a vertex normal cone is the union of its facets shared with neighboring cones. Thus, at a cost with a unique optimizer, its margin equals its angular distance to the boundary of its normal cone. At a tied cost, this distance is zero for every optimal vertex. 

Suppose that $\tilde{w}(x) \in W^\star(c_a)$ and $c \in \mathcal{E}$. Then, we have $\tilde{w}(x) \neq w_0$. This implies that $\cone(w_0)$ and $\cone(\tilde{w}(x))$ have disjoint interiors. We consider a shortest spherical path from $\frac{c}{\|c\|}$ to $\frac{c_a}{\|c_a\|}$. If $c$ has a unique optimizer, the initial segment before leaving the interior of $\cone(w_0)$ has length at least $\psi(x)$. If $c_a$ has a unique optimizer, the final segment after its last entry into the interior of $\cone(\tilde{w}(x))$ has length at least $\phi(x)$. At a tied endpoint, we take the corresponding segment to have length zero, equal to its margin. The two segments do not overlap, giving
\begin{equation*}
L_\textnormal{pred}(P^a,P^{DM};x) = \angle(c_a,c) \geq \psi(x)+\phi(x).
\end{equation*}
In particular, the loss is at least $\epsilon$ when $\phi(x) \geq \epsilon$.

Next, we consider the sufficient condition for fitting. If $\phi(x)=0$, its strict inequality cannot hold since the
optimality gap is nonnegative. Otherwise, $w_a$ is the unique optimizer under $c_a$. For every $v \in V\setminus \{w_a\}$, we have $\cone(w_a) \subseteq \{u \in \R^n: u^\top(v-w_a) \geq 0\}$. A shortest spherical path from $\frac{c_a}{\|c_a\|}$ to the hyper-plane $u^\top(v-w_a)=0$ must reach the cone boundary no later than its endpoint. Thus, we have
\begin{equation*} 
0 < \phi(x) \leq \arcsin\left(\tfrac{c_a^\top(v-w_a)}{\|c_a\|\|v-w_a\|}\right) \leq \tfrac{\pi}{2}.
\end{equation*}
This implies 
\begin{equation*}
c_a^\top(v-w_a) \geq \|c_a\|\|v-w_a\|\sin(\phi(x)) \geq \|c_a\|\ell_e\sin(\phi(x)).
\end{equation*}
Therefore, we conclude that $c_a^\top(\tilde{w}(x)-w_a)<\|c_a\|\ell_e\sin(\phi(x))$ forces $\tilde{w}(x)=w_a\in W^\star(c_a)$.

For the expected bound, the cache can equivalently be generated by assigning an independent uniformly chosen member to every possible candidate set before queries begin. For each fixed $x$, the set $\mathcal{W}_\epsilon(x)$ is independent of these assignments and contains $w^\star(P^{DM}x)$. Averaging over the cache randomness yields
\begin{equation*}
\mathbb{P}\{\tilde{w}(x) \neq w^\star(P^{DM}x)\} = 1-\tfrac{1}{|\mathcal{W}_\epsilon(x)|}.
\end{equation*}
Under the stated assumption, the pointwise bound is at least $\psi(x)$ on this event and is nonnegative otherwise. Averaging over the defense and attacker randomness and an independent $x\sim Q$ therefore yields
\begin{equation*} 
\mathbb{E}[L_\textnormal{pred}(P^a,P^{DM};x)] \geq \mathbb{E}_{x \sim Q}\left[\psi(x)\left(1-\tfrac{1}{|\mathcal{W}_\epsilon(x)|}\right)\right] = B_\epsilon(Q).
\end{equation*}
Finally, $\mathcal{W}_\epsilon(x)$ is nondecreasing in $\epsilon$ under set inclusion, including at $\epsilon=0$, while $\psi(x) \geq 0$ is independent of $\epsilon$. Therefore, we conclude that $B_\epsilon(Q)$ is nondecreasing in $\epsilon$.
\end{proof}

\begin{proof}[Proof of Proposition \ref{prop:1-dim-suff-condition}]
Fix a full-row-rank matrix $P^a$ and $P \in \mathcal{P}_c(P^a)$. For each facet $i$, we choose $x_i \in\R^d$ with $P^a x_i=r_i$. If $z \in \ker(P^a)$, we have $P^a(x_i+z)=r_i$. This implies $Px_i, P(x_i+z) \in \operatorname{span}(r_i)$ and $Pz \in \operatorname{span}(r_i)$. Since the normals include two independent directions, we have $Pz=0$. This implies $\ker(P^a) \subseteq \ker(P)$ and the full row rank
of $P^a$ gives $P=BP^a$ for a linear map $B: \R^n \to \R^n$.

For each $i$, we have $Br_i = Px_i \in \operatorname{span}(r_i)$ or equivalently $Ba_i=\lambda_i a_i$ for some $\lambda_i \in \R$. Applying $B$ to the relation in Assumption~\ref{assumption:main} yields $\sum_{j=1}^n \beta_j(\lambda_{k_j}-\lambda_{k_{n+1}})a_{k_j}=0$. Linear independence and $\beta_j \neq 0$ imply $\lambda_{k_j}=\lambda_{k_{n+1}}$ for every $j=1,\ldots,n$. Thus, we have $B=\gamma I_n$ and $P=\gamma P^a$ for some $\gamma \in \R$. Conversely, every real multiple of $P^a$ belongs to the candidate space. Thus, we have $\mathcal{P}_c(P^a)=\operatorname{span}\{P^a\}$ and its dimension is $1$.  
\end{proof}

\begin{proof}[Proof of Corollary \ref{coro:calibrate-condition-DM}]
Suppose that $P^a \neq 0$ satisfies $w^\star(P^{DM}x) \in W^\star(P^a x)$ for every $x \notin \ker(P^{DM})$. Then, the proof of Theorem~\ref{thm:loss-cali-cond} gives a linear map $B:\R^n \to \R^n$ satisfying $P^a=BP^{DM}$ and $Br_i \in \{\alpha r_i: \alpha \geq 0\}$ for every $i$. Thus, we have $P^a\bar{x} \in \operatorname{span}(r_i)$ whenever $P^{DM}\bar{x}=r_i$. This implies $P^a \in \mathcal{P}_c(P^{DM})$ and $P^a=\gamma P^{DM}$ for some $\gamma \in \R$. For any $\bar{x}$ with $P^{DM}\bar{x}=r_i$, we have $\gamma r_i=P^a\bar{x} = Br_i \in \{\alpha r_i: \alpha \geq 0\}$. Since $P^a \neq 0$, this implies $\gamma>0$. Thus, every inverse feasibility loss calibrates the prediction loss. 
\end{proof}

\begin{lemma}\label{lem:sample-error-edge}
For each $p \geq 1$, we let $U_p \in \R^{n\times d}$ be a full-row-rank terminal output of Algorithm~\ref{algo:attack} with parameters $M_p \geq 1$ and $\Delta_p > 0$, normalized so that $\|U_p\|_F=1$. Fix a nonempty set $J \subseteq \{1,\ldots,m\}$. Suppose its final sampling call uses the construction in
Algorithm~\ref{algo:edge-sampling} and retains every boundary and near-boundary query with $i \in J$. Denote the boundary representatives and near-boundary queries in that call by $\bar{x}_{i0j,p}$ and $\tilde{x}_{ikj,p}$, respectively, where $j=1,\ldots,d-n+1$ and $k \in \mathcal{C}_i$. In particular, $U_p\bar{ x}_{i0j,p}=r_i$. Assume that every boundary representative used to form these queries, including auxiliary representatives outside $J$, has norm at most a common constant $R>0$. If $M_p \rightarrow +\infty$ and $\Delta_p(M_p+K-1)<\eta_S$ for all sufficiently large $p$, we have
\begin{equation*}
\inf_{\alpha \geq 0} \|P^{DM}\bar{x}_{i0j,p}-\alpha r_i\| \rightarrow 0 \textnormal{ for all } i\in J,\ j=1,\ldots,d-n+1, 
\end{equation*}
where $K$ and $\eta_S$ are defined in Appendix~\ref{sec:app:sampling}. 
\end{lemma}
\begin{proof}
Fix $i \in J$, $j \in \{1,\ldots,d-n+1\}$, and $k \in \mathcal{C}_i$. Using the estimate in Appendix~\ref{sec:app:sampling} and the stated stopping condition, for sufficiently large $p$, we have
\begin{equation*} 
(U_p\tilde{x}_{ikj,p})^\top(v-w_k) \geq \tfrac{\eta_S}{M_p+K-1} > \Delta_p \textnormal{ for every } v \in V \setminus \{w_k\}.
\end{equation*}
Since this query is retained in the dataset on which $U_p$ satisfies the stopping test, its returned vertex must be $w_k$. Thus, we have $P^{DM}\tilde{x}_{ikj,p} \in C_k$. For $M_p>K-1$, we have
\begin{equation*} 
\|\tilde{x}_{ikj,p}-\bar{x}_{i0j,p}\| \leq \tfrac{2R(K-1)}{M_p-(K-1)} \rightarrow 0, 
\end{equation*}
which implies 
\begin{equation*} 
\|P^{DM}\tilde{x}_{ikj,p}-P^{DM}\bar{x}_{i0j,p}\| \leq \|P^{DM}\|_F\|\tilde{x}_{ikj,p}-\bar{x}_{i0j,p}\| \rightarrow 0.
\end{equation*}
The sequence $\{P^{DM}\bar{x}_{i0j,p}\}_{p \geq 1}$ is bounded. Since each $C_k$ is closed, every subsequential limit lies in $\cap_{k \in \mathcal{C}_i} C_k = \{\alpha r_i: \alpha \geq 0\}$. If $\inf_{\alpha \geq 0} \|P^{DM}\bar{x}_{i0j,p}-\alpha r_i\|$ did not tend to zero, a bounded subsequence staying a positive distance from this closed ray would have a limit in the ray, yielding a contradiction. 
\end{proof}

\begin{lemma}\label{lem:sample-global-error}
Let $J \subseteq \{1,\ldots,m\}$ be such that $\{a_i:i\in J\}$ contains $n+1$ normals satisfying Assumption~\ref{assumption:main}. Let $\{U_p\}_{p \geq 1} \subseteq \R^{n\times d}$ consist of
full-row-rank matrices with $\|U_p\|_F=1$. For each $i \in J$ and $p \geq 1$, we choose $\bar{x}_{i0j,p} \in\R^d$ satisfying $U_p\bar{x}_{i0j,p}=r_i$ for $j=1,\ldots,d-n+1$, and set $X_{i,p}=[\bar{x}_{i01,p},\ldots,\bar{x}_{i0,d-n+1,p}]$. Suppose there are constants $R,\kappa>0$, independent of $p$, such that $\|\bar{x}_{i0j,p}\| \leq R$ and $\sigma_{\min}(X_{i,p}) \geq \kappa$ for every $i \in J$ and $j$, where $\sigma_{\min}$ is the smallest of the $d-n+1$ singular values. If $\inf_{\alpha \geq 0} \|P^{DM}\bar{x}_{i0j,p}-\alpha r_i\| \rightarrow 0$ for all $i \in J$ and $j=1,\ldots,d-n+1$, we have $U_p \rightarrow \tfrac{P^{DM}}{\|P^{DM}\|_F}$ and $\angle(U_p x,P^{DM}x) \rightarrow 0$ for every $x \notin \ker(P^{DM})$. 
\end{lemma}
\begin{proof}
Since $\|U_p\|_F=1$, it suffices to identify every subsequential limit. We take a subsequence with $U_p \to  U_\star$. Boundedness of the finitely many representatives gives a further subsequence, denoted by the same indices, such that $\bar{x}_{i0j,p} \rightarrow \bar{x}_{i0j,\star}$ for all $i\in J$ and $j=1,\ldots,d-n+1$. 

We define $X_{i,\star}=[\bar{x}_{i01,\star},\ldots,\bar{x}_{i0,d-n+1,\star}]$. Then, $X_{i,p} \to X_{i,\star}$, and continuity gives $\|U_\star\|_F=1$, $U_\star\bar{x}_{i0j,\star}=r_i$ and $\sigma_{\min}(X_{i,\star}) \geq \kappa$. The image of $U_\star$ contains the selected $r_i$, including $n$ independent directions. Thus, $U_\star$ has full row rank. For each $i \in J$, the columns of $X_{i,\star}$ lie in a ($d-n+1$)-dimensional space $\{x \in \R^d: U_\star x \in \operatorname{span}(r_i)\}$. There are $d-n+1$ such columns, and the conditioning bound makes them independent, so they form a basis of this space. The assumed convergence gives $P^{DM}\bar{x}_{i0j,\star} \in \{\alpha r_i: \alpha \geq 0\}$. By linearity, we have $P^{DM}x \in \operatorname{span}(r_i)$ whenever $U_\star x \in \operatorname{span}(r_i)$, and in particular, whenever $U_\star x=r_i$. 

Since $J$ contains $n+1$ normals satisfying Assumption~\ref{assumption:main}, the argument in the proof of Proposition~\ref{prop:1-dim-suff-condition}, restricted to $i\in J$, gives $P^{DM}=\gamma U_\star$ for some $\gamma \in \R$. Since $\gamma r_i=P^{DM}\bar{x}_{i0j,\star} \in \{\alpha r_i: \alpha \geq 0\}$ and $P^{DM} \neq 0$, we have $\gamma>0$. Normalization then gives $\gamma=\|P^{DM}\|_F$ and $U_\star=\frac{P^{DM}}{\|P^{DM}\|_F}$. Every subsequential limit is the same, proving convergence
of the full sequence. For every $x \notin \ker(P^{\mathrm{DM}})$, we have $U_p x \rightarrow \frac{P^{DM}x}{\|P^{DM}\|_F} \neq 0$. Continuity of angular distance therefore gives the desired convergence of the prediction loss.
\end{proof}

\begin{proof}[Proof of Theorem \ref{thm:sample-global-error}]
Under the stated assumptions, Proposition~\ref{prop:adv-algo-stop} guarantees that every run terminates after at most $\left(1+\frac{2R L_e}{\Delta}\right)^{nd}$ inverse optimization updates. We take any sequence $\Delta_p \to 0$ in the specified family and set $M_p=M(\Delta_p)$ and $U_p=P^a_{M_p,\Delta_p}$. For each $p$, we let $\bar{x}_{i0j,p}$ denote the boundary representatives in the final sampling call associated with $U_p$, and set $X_{i,p}=[\bar{x}_{i01,p},\ldots,\bar{x}_{i0,d-n+1,p}]$. This call is included in the dataset used by the stopping test. Every boundary representative is queried and hence has norm at most $R$. Since $M_p \to +\infty$ and the stopping condition holds for sufficiently large $p$, Lemma~\ref{lem:sample-error-edge} applies with $J=\{1,\ldots,m\}$ and gives
\begin{equation*}
\inf_{\alpha \geq 0} \|P^{DM}\bar{x}_{i0j,p}-\alpha r_i\| \rightarrow 0 \textnormal{ for all } i=1,\ldots,m,\ j=1,\ldots,d-n+1.
\end{equation*}
The bound $\sigma_{\min}(X_{i,p}) \geq \kappa$ then allows us to apply Lemma~\ref{lem:sample-global-error}, yielding $P^a_{M_p,\Delta_p} \rightarrow \frac{P^{DM}}{\|P^{DM}\|_F}$. Since the sequence $\Delta_p \to 0$ was arbitrary, the matrix convergence holds for the full family. For every $x \notin \ker(P^{DM})$, the same lemma gives $L_\textnormal{pred}(P^a_{M(\Delta), \Delta}, P^{DM};x) \rightarrow 0$.
\end{proof}

\begin{proof}[Proof of Lemma \ref{lem:1-dim-suff-condition-restrict-region}]
Suppose that $P^a \neq 0$ satisfies the stated consistency condition on $\mathcal{X} \setminus \ker(P^{DM})$. Then, for each $i\in J$, we fix $\bar{x}_i \in \operatorname{int}(\mathcal{X})$ and $\alpha_i>0$ from the assumption, so that $P^{DM}\bar{x}_i=\alpha_i r_i$. We fix $i\in J$ and any $x \in\operatorname{int}(\mathcal{X})$ with $P^{DM}x=\alpha_i r_i$. For each $k \in \mathcal{C}_i$, we choose $h_k \in \R^d$ with $P^{DM}h_k \in \operatorname{int}(C_k)$, using the full row rank of $P^{DM}$. For sufficiently small $t>0$, we have $x+t h_k \in \mathcal{X}$ and $P^{DM}(x+t h_k)=\alpha_i r_i+tP^{DM}h_k \in \operatorname{int}(C_k)$. Consistency gives $P^a(x+t h_k) \in C_k$. Letting $t \downarrow 0$ and using closedness yields $P^a x \in \cap_{k \in \mathcal{C}_i} C_k = \{\alpha r_i: \alpha \geq 0\}$. 

We fix $z \in \ker(P^{DM})$. For each $i \in J$ and sufficiently small nonzero $t$, both $\bar{x}_i$ and $\bar{x}_i+tz$ lie in $\operatorname{int}(\mathcal{X})$ and have deployed cost $\alpha_i r_i$. The preceding conclusion gives $P^a\bar{x}_i, P^a(\bar{x}_i+tz) \in \{\alpha r_i: \alpha \geq 0\}$ and $P^a z \in\operatorname{span}(r_i)$. Since the selected normals include two independent directions, we have $P^a z=0$. This implies $\ker(P^{DM}) \subseteq \ker(P^a)$ and $P^a=BP^{DM}$ for a linear map $B: \R^n \to \R^n$. Evaluating at $\bar{x}_i$ gives $\alpha_i Br_i=P^a\bar{x}_i \in \{\alpha r_i: \alpha \geq 0\}$. Thus, we have $Br_i=\lambda_i r_i$ for some $\lambda_i \geq 0$ for each $i\in J$. Applying the eigenvalue argument from the proof of Theorem~\ref{thm:loss-cali-cond} to the selected $n+1$ normals gives $B=\gamma I_n$ with $\gamma \geq 0$. Since $P^a \neq 0$, we have $\gamma>0$ and $P^a=\gamma P^{DM}$.
\end{proof}

\begin{proof}[Proof of Lemma \ref{lem:restricted-boundary-sampling-convergence}]
Under the stated assumptions, Proposition~\ref{prop:adv-algo-stop} guarantees that every run terminates after at most $\left(1+\frac{2R L_e}{\Delta}\right)^{nd}$ inverse optimization updates. We take any sequence $\Delta_p \to 0$ in the specified family and set $M_p = M(\Delta_p)$ and $U_p=P^a_{M_p,\Delta_p}$. For each $p$, we let $\bar{x}_{i0j,p}$ denote the boundary representatives in the final sampling call associated with $U_p$, and set $X_{i,p}=[\bar{x}_{i01,p},\ldots,\bar{x}_{i0,d-n+1,p}]$ for all $i\in J$.
Every query in each selected block $i\in J$ is retained in the dataset used by the stopping test. All representatives used to form these queries, including those outside $J$, have norm at most $R$. Lemma~\ref{lem:sample-error-edge} therefore gives
\begin{equation*} 
\inf_{\alpha \geq 0} \|P^{DM}\bar{x}_{i0j,p}-\alpha r_i\| \rightarrow 0 \textnormal{ for all } i\in J,\ j=1,\ldots,d-n+1.
\end{equation*}
The selected facets contain $n+1$ normals satisfying Assumption~\ref{assumption:main}, and $\sigma_{\min}(X_{i,p}) \geq \kappa$ for every $i\in J$. Lemma~\ref{lem:sample-global-error} yields $U_p \rightarrow \frac{P^{DM}}{\|P^{DM}\|_F}$. Since the sequence $\Delta_p \rightarrow 0$ was arbitrary, the matrix convergence holds for the full family. For every $x \notin \ker(P^{DM})$, the same lemma gives $L_\textnormal{pred}(P^a_{M(\Delta), \Delta}, P^{DM};x) \rightarrow 0$.
\end{proof}

\end{document}